\documentclass[journal,onecolumn,12pt]{IEEEtran}

\pdfoutput=1
\usepackage{cite}
\usepackage[dvipsnames]{xcolor}

\ifCLASSINFOpdf
  \usepackage[pdftex]{graphicx}
  \graphicspath{{images/}}
  \DeclareGraphicsExtensions{.pdf,.jpeg,.png}
\else
  \usepackage[dvips]{graphicx}
  \graphicspath{{images/}}
  \DeclareGraphicsExtensions{.eps}
\fi

\usepackage{amsmath}
\usepackage{array}

\ifCLASSOPTIONcompsoc
  \usepackage[caption=false,font=normalsize,labelfont=sf,textfont=sf]{subfig}
\else
  \usepackage[caption=false,font=footnotesize]{subfig}
\fi

\usepackage{pgfplots}
\usepackage{float}
\usepackage{stfloats}

\usepackage{url}

\usepackage{upgreek}
\usepackage{framed}

\newcommand\citet\cite

\newlength{\fwidth}
\newcommand{\labelsize}{\footnotesize}
\newcommand{\ticksize}{\scriptsize}
\def\xlabeldist{0.1}
\def\ylabeldist{0.06}

\usepackage{hyperref}
\usepackage[capitalise]{cleveref}

\usepgfplotslibrary{external}
\tikzset{>=stealth}

\begin{document}
	
\IEEEspecialpapernotice{This is a submitted preprint version of a published paper, please refer to the final peer reviewed and published version at \url{https://doi.org/10.1002/rob.21764}. Additionally, please cite with\\ \textbf{Pijnaker Hordijk B.J., Scheper K.Y.W., de Croon GCHE. Vertical Landing for Micro Air Vehicles using Event-based Optical Flow. J Field Robotics. 2017;00:1-22. https://doi.org/10.1002/rob.21764}}

%
\title{Vertical Landing for Micro Air Vehicles using Event-Based Optical Flow}
%
%
%

\author{\IEEEauthorblockN{Bas J. Pijnacker Hordijk}\\
\IEEEauthorblockA{Faculty of Aerospace Engineering\\
Delft University of Technology\\
2629HS Delft, The Netherlands\\ \vspace{1cm}}
\and
\IEEEauthorblockN{Kirk Y.W. Scheper\IEEEauthorrefmark{1}\thanks{\IEEEauthorrefmark{1} Corresponding author.}}\\
\IEEEauthorblockA{Micro Air Vehicle Laboratory\\
Faculty of Aerospace Engineering\\
Delft University of Technology\\
2629HS Delft, The Netherlands\\
k.y.w.scheper@tudelft.nl\\ \vspace{1cm}}
\and
\IEEEauthorblockN{Guido C.H.E. De Croon}\\
\IEEEauthorblockA{Micro Air Vehicle Laboratory\\
Faculty of Aerospace Engineering\\
Delft University of Technology\\
2629HS Delft, The Netherlands\\
g.c.h.e.decroon@tudelft.nl}}

\maketitle

\begin{abstract}

Small flying robots can perform landing maneuvers using bio-inspired optical flow by maintaining a constant divergence. However, optical flow is typically estimated from frame sequences recorded by standard miniature cameras. This requires processing full images on-board, limiting the update rate of divergence measurements, and thus the speed of the control loop and the robot. Event-based cameras overcome these limitations by only measuring pixel-level brightness changes at microsecond temporal accuracy, hence providing an efficient mechanism for optical flow estimation. This paper presents, to the best of our knowledge, the first work integrating event-based optical flow estimation into the control loop of a flying robot. We extend an existing 'local plane fitting' algorithm to obtain an improved and more computationally efficient optical flow estimation method, valid for a wide range of optical flow velocities. This method is validated for real event sequences. In addition, a method for estimating the divergence from event-based optical flow is introduced, which accounts for the aperture problem. The developed algorithms are implemented in a constant divergence landing controller on-board of a quadrotor. Experiments show that, using event-based optical flow, accurate divergence estimates can be obtained over a wide range of speeds. This enables the quadrotor to perform very fast landing maneuvers.





\end{abstract}


%
\IEEEpeerreviewmaketitle

\section{Introduction}

Rapid advances in micro-electronics catalyzed the development of tiny flying robots \cite{Floreano2015}, formally referred to as Micro Air Vehicles (MAVs). Due to their size and agility, MAVs have the potential to perform activities in confined and cluttered environments. However, achieving autonomous flight with very small MAVs (for example, the 20-gram DelFly Explorer \cite{DeWagter2014}) is a significant challenge due to strict weight and power limitations for on-board equipment. The speed of cutting edge autonomous MAV navigation pales in comparison to their natural counterparts, the insect.

The main sensor system of most insects uses some form of visual light to perceive the environment around them. Visual navigation in flying insects is primarily based on optical flow, the apparent motion of brightness patterns perceived by an observer due to relative motion with respect to the environment \cite{Gibson1979}. In essence, optical flow provides information on the ratio of velocity to distance, such that the actual metric distance to the environment is not directly available. Instead, flying insects navigate based on certain visual observables extracted from the optical flow field that relate to ego-motion. Honeybees were seen to control their descent during landings based on the \emph{divergence} of the optical flow field perceived from the ground \cite{Baird2013}. When looking down, flow field divergence conveys the vertical velocity scaled by the height. By maintaining a constant divergence in downward motion, an observer approaches the ground while exponentially decreasing its downward speed. For flying robots capable of vertical landing, this is an interesting strategy. This application has been explored in several experiments with rotorcraft MAVs \cite{Herisse2008,Herisse2012,Ho2016,DeCroon2016}.

While such optical flow based navigation strategies are bio-inspired, most visual sensors employed for measuring optical flow differ significantly from their natural counterparts. Commonly used miniature cameras operate in a \emph{frame-based} manner: full frames are obtained by periodically measuring brightness at all pixels. This is a relatively inefficient process for motion perception, since the information output rate is independent of the actual dynamics in the scene. Static parts of a frame are recorded as well as changing parts, even though only the latter are relevant for motion perception. Therefore, follow-up processing of a full frame is necessary, which at present still requires significant processing. In addition, fast inter-frame displacements lead to motion blur, which limits the accuracy of resulting optical flow estimates or requires complex algorithms to account for this. These characteristics are undesirable for optical flow measurement on board miniature flying robots, which are subject to strict computational limits and exhibit fast dynamics. 

In contrast, biological vision systems, such as the compound eyes of insects, employ an \emph{event-driven} mechanism; they measure changes in the perceived scene at the moment of detection \cite{Linares-barranco2014}. Several researchers have attempted to minic the sensory system in insects in order to measure optical flow. For example, in \citet{Ruffier2014} a tethered rotorcraft MAV was equipped with a 2-photodetector neuromorphic chip for measuring translational optical flow. In \citet{Floreano2013} a miniature curved compound eye design, inspired by the fruit fly Drosophila, was presented.

In particular, \emph{event-based cameras} are a promising class of sensors for optical flow perception. When a pixel of an event-based camera measures a relative increase or decrease in brightness, it registers an event, mapping its pixel location to the timestamp and sign of the change. This timestamp is obtained with microsecond resolution and latencies in the range of 3 to 15 $\upmu$s. In addition, event-based pixel architectures enable intrascene dynamic ranges exceeding 120 dB \cite{Yang2015}. These characteristics are highly desirable in robotic visual navigation. Experiments using event-based cameras demonstrated high performance of visual control systems through low-latency state updates, efficient data processing, and operation over a wide range of illumination conditions \cite{Conradt2009,Delbruck2013}.

This novel approach to visual sensing is in general incompatible with state-of-the-art computer vision algorithms for estimating optical flow, due to the lack of absolute brightness measurements. Therefore, several event-based methods for optical flow estimation \cite{Benosman2012,Benosman2014, Barranco2014, Brosch2015, Bardow2016} as well as benchmarking datasets \cite{Barranco2016, Ruckauer2016} have been developed. Of the existing techniques, the local plane fitting approach by \citet{Benosman2014} is the most promising based on its application to estimating time-to-contact (the reciprocal of flow field divergence) in simple scenes \cite{Clady2014} and good results in \citet{Ruckauer2016}. However, until now, no study has discussed an implementation of event-based optical flow into an optical flow based control loop of an MAV. 

This paper contains \emph{three main contributions}. First, a novel method for estimating event-based optical flow inspired by \citet{Benosman2014} is introduced. Its applicability is extended to a wider range of velocities, while improving computational efficiency. Second, a method for incorporating event-based optical flow into visual estimation of divergence is proposed, which accounts for the aperture problem that occurs in most existing event-based optical flow methods. Third, the proposed algorithms for event-based optical flow divergence estimation are incorporated in a constant divergence landing controller on-board of a quadrotor. To the best of the authors' knowledge, this paper presents the first functional event-based visual controller on-board of an MAV.
 
We begin by discussing related work concerning landing using optical flow, event-based vision, and optical flow estimation in \cref{sec:related}. Then, \cref{sec:model} introduces the mathematical models describing the relations between the ego-motion of the MAV, optical flow, and visual observables used in this work. In \cref{sec:eof_estimation} the estimation method for event-based optical flow is described and evaluated. Subsequently, \cref{sec:landing_eof} introduces the approach to estimating visual observables from event-based optical flow, followed by an assessment of its performance in combination with the optical flow method. Afterwards, \cref{sec:results} presents flight test results of the full estimation pipeline during constant divergence landing maneuvers. Lastly, the main conclusions and recommendations for future work are presented in \cref{sec:conclusion}.


\section{Related Work}
\label{sec:related}

This section introduces the fundamental concepts and previous contributions relevant for this work. First, \cref{sec:related_landing} discusses bio-inspired landing strategies involving optical flow and associated research involving MAVs. Second, the concept of event-based cameras is described in \cref{sec:related_event_cams}. Third, an overview of existing approaches to optical flow estimation is provided in \cref{sec:related_optical_flow}, including both frame-based camera applications and recently developed event-based techniques.

\subsection{Landing Using Optical Flow} 
\label{sec:related_landing}
Although optical flow does not by itself provide metric scale to motion, information from optical flow fields is useful for several navigation tasks, including landing. Simple bio-inspired strategies were proposed in the past decades that utilize the visual observables in the optical flow field perceived from the ground. Such strategies form a lightweight alternative for visual estimation of three-dimensional structure, ego-motion, and relative pose, which can be performed through visual Simultaneous Localization And Mapping (SLAM) \cite{Davison2007a} or visual odometry \cite{Nister2004}. These techniques have become increasingly efficient over the last few years \cite{Forster2014,Engel2014}, yet still require processing and maintaining large amounts of measurement data. This strongly contrasts with optical flow based techniques, in which all information required for navigation is contained in a small number of visual observables.

The visual observables related to horizontal motion above ground are the ventral flows $\omega_x$, $\omega_y$, referring to the average flows along the $x$ and $y$ image axes. In several experiments with a tethered MAV \cite{Ruffier2014,Expert2015}, the authors mimicked navigation strategies seen in insects, which have been observed to follow terrain and land using ventral flow \cite{Srinivasan1996}. By maintaining a constant ratio between forward motion and height and at the same time slowing down, they perform smooth landings. Ventral flows may also be used for hover stabilization to augment visual vertical control \cite{Alkowatly2015}.

In recent aerial robotic applications, mainly visual observables based on vertical motion were applied, allowing control of vertical dynamics independent of horizontal motion. One of these observables is flow field divergence $D$, i.e. the ratio of vertical velocity to height. Its reciprocal is the time-to-contact to the ground $\tau$. Similar to ventral flows, divergence was seen to guide docking and landing motion in biology. In \cite{Baird2013} honeybees were seen to keep $D$ constant. Hence, velocity is decreased exponentially, ensuring a smooth touchdown. Other strategies for vertical landing exist based on $\tau$, such as the constantly decreasing $\tau$ strategy observed in braking human drivers \cite{Lee1976}. This strategy provides more control over the landing trajectory \cite{Alkowatly2015}, though it involves more parameters.

In practical control systems, a constant divergence approach suffers from instability as height decreases, due to self-induced oscillations. In \citet{DeCroon2016} it was shown that a relation exists between the employed divergence controller gain and the height at which oscillations occur. A main insight then was that a drone could detect its own oscillations, and in this way determine its height. This strategy can be employed to trigger a separate final touchdown phase, or to continuously measure height by landing at a near-unstable control gain. The finding in \citet{DeCroon2016} also contains an important key to high-performance optical flow divergence landings. This was used in \citet{Ho2016a} to develop an adaptive gain controller, which detects the height at the start of a landing maneuver, sets initial controller gains based on the height, and lands while exponentially reducing the gains. Although the landings performed were quite fast compared to landings in the literature ($D=0.3$ compared to a typical $D=0.05$ \cite{Herisse2012}), the speed of the landings in \citet{Ho2016a} are still quite limited by the standard cameras available on the used AR drone 2.0.

\subsection{Event-Based Cameras}
\label{sec:related_event_cams}
Inspired by the workings of biological retinas, event-based cameras rely on a sensing mechanism that fundamentally differs from their frame-based counterparts. In frame-based cameras the pixel values are measured at fixed time intervals to produce a sequence of images. In event-based cameras, on the other hand, pixel activity is driven by light intensity changes. Whenever a pixel measures a local change, it produces a \emph{event}. Specifically, this occurs when the pixel's logarithmic intensity measurement $I(x,y,t)$ (at pixel location $(x,y)$ and timestamp $t$) increases or decreases beyond a threshold $C$:

\begin{equation}
\label{eq:event_threshold}
\lvert\Delta\left(\log I\left(x,y,t\right)\right)\rvert > C
\end{equation}

\begin{figure}[!htpb]
	\centering
	\includegraphics[width=.65\linewidth]{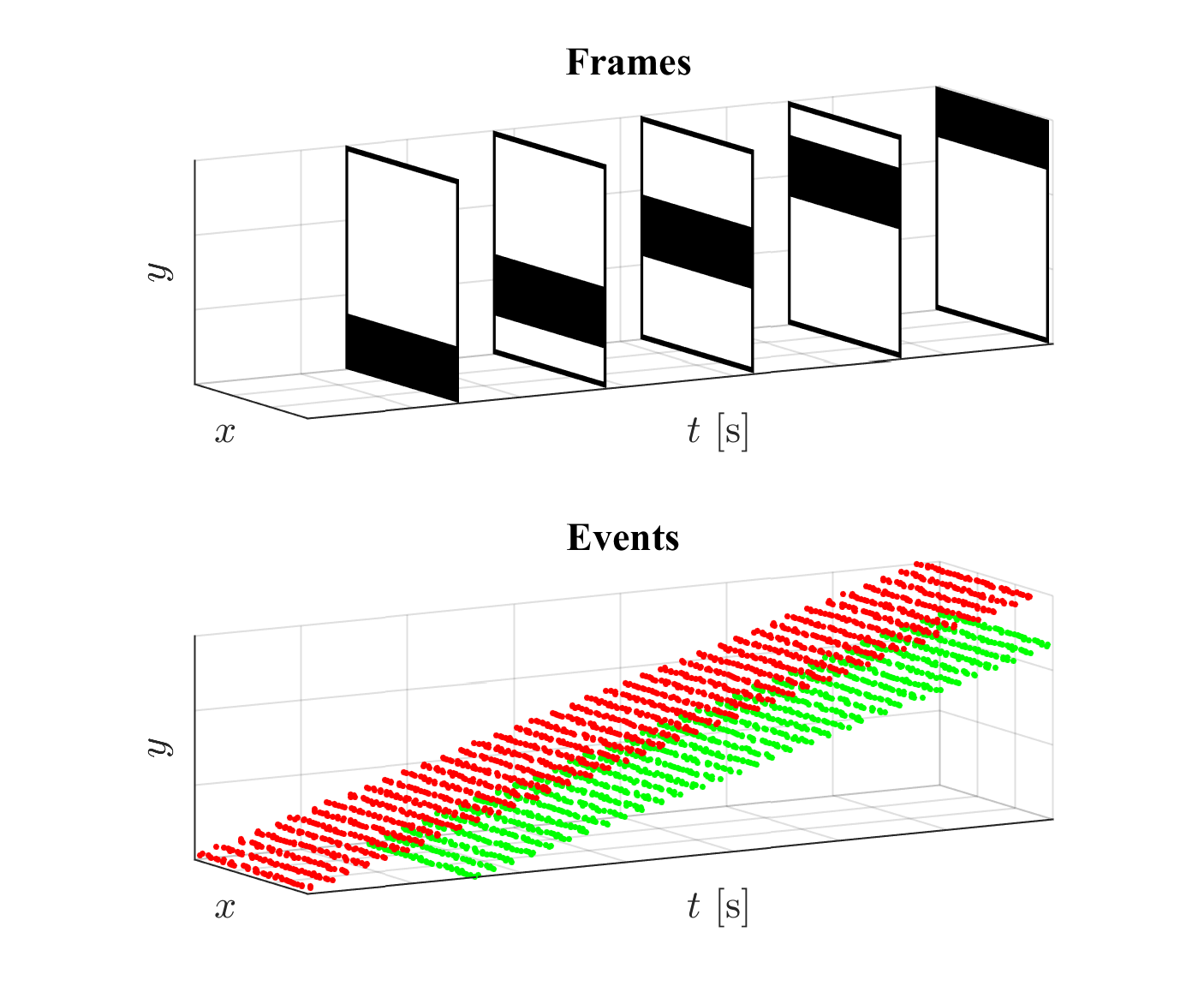}
	\caption{Frame-based and event-based visual output generated from a simple synthetic scene, in which a black horizontal bar moves upward. The events are visualized as points in space-time, hence showing the trajectory of the leading and trailing edges of the black bar. Events with positive polarity are highlighted in green; those with negative polarity are marked in red.}
	\label{fig:events_frames}
\end{figure}

Events are encoded according to an Address-Event Representation (AER) \cite{Lichtsteiner2008}, which consists of event information encoded by an address and the timestamp of detection. Typically, an event encodes the pixel position $(x,y)$, the timestamp $t$, and the polarity $P\in\lbrace-1,1\rbrace$, which indicates the sign of the intensity change. A visualization of a basic stream of events, in comparison to an equivalent set of frames, is shown in \cref{fig:events_frames}.

\begin{figure}[!t]
	\centering
	\includegraphics[width=0.3\linewidth]{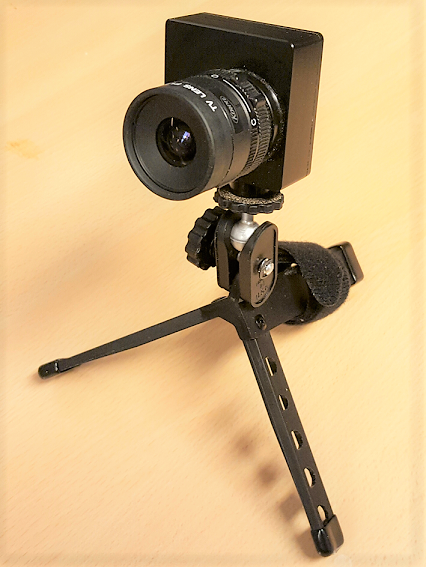}
	\caption{Picture of the event-based camera employed in this work, the DVS.}
	\label{fig:dvs}
\end{figure}

The sensor used in this work is the Dynamic Vision Sensor (DVS) - specifically, the DVS128 - which is the first commercially available event-based camera \cite{Cho2015}. It features a 128x128 pixel grid operating at an intrascene dynamic range of 120 dB, measuring events at 1 $\upmu$s timing resolution with a latency of 15 $\upmu$s \cite{Lichtsteiner2008}. A picture of the DVS is shown in \cref{fig:dvs}. Since the availability of the DVS, other event-based cameras have been developed. Most notable are the Asynchronous Time-based Image Sensor (ATIS) \cite{Posch2011}, which measures absolute intensity as well as polarity for each event, and the Dynamic and Active pixel Vision Sensor (DAVIS) \cite{Brandli2014}, whose pixels record events as well as full frames. Interesting in the context of this work is the 2.2 gram micro embedded DVS (meDVS) \cite{Conradt2015}, which is highly suitable for on-board MAV applications.

Event-based cameras have several interesting applications for robotic navigation. Initial work has been performed on visual SLAM with event-based cameras \cite{Weikersdorfer2013,Weikersdorfer2014}. In \citet{Mueggler2014} a pose estimation algorithm based on line tracking is applied to a quadrotor, enabling it to perform aggressive maneuvers . Some studies demonstrate the ability to simultaneously reconstruct intensity maps and relative pose \cite{Kim2014} and, more recently, three-dimensional structure \cite{Kim2016}. Others aim at combining the benefits of event-based and frame-based vision using the DAVIS. For example, the method presented in \citet{Kueng2016} uses frames to identify visual features and events to track their position in high-speed motion, in order to perform visual odometry.

\subsection{Optical Flow Estimation}
\label{sec:related_optical_flow}
In the following, we discuss available techniques for estimating optical flow. Many recent visual navigation experiments, in particular those with commercially available quadrotors, employ standard frame-based cameras, in combination with follow-up processing algorithms. Others employ off-the-shelf optical mouse sensors or specialized neuromorphic optical flow sensors, which directly yield translational optical flow output. For event-based cameras, several techniques have recently been developed, yet these have seen limited applications in robotic navigation.

\subsubsection{Estimation from Frame Sequences}
At present, a wide range of optical flow estimation techniques is available for frame-based cameras. Most of these algorithms derive from the brightness constancy assumption, which states that, when a pixel flows from one frame to another, its intensity $I$ is conserved \cite{Baker2011}. This assumption leads to the well-known optical flow constraint:

\begin{equation}
\label{eq:optical_flow_constraint}
I_x(x,y) u + I_y(x,y) v = -I_t(x,y)
\end{equation}

where $x$ and $y$ are the pixel position and $u$ and $v$ denote the unknown optical flow components in pixels per second. The partial derivatives $I_x$, $I_y$, and $I_t$ are obtained from two sequential frames. Since this equation provides two unknown components, a second constraint is necessary to obtain optical flow. Many recent methods aim at providing a dense optical flow field estimate, where optical flow is estimated for any pixel for the frame. In this case, a global cost function minimization is performed, in which a second constraint is provided by prior knowledge. An example of such a constraint is the requirement of smoothness in the flow field in the well-known Horn-Schunck technique \cite{Horn1981}. Recent dense optical flow algorithms provide accurate results for complex scenes, but at the cost of high computation times \cite{Baker2011}.

In recent real-time robotic applications, the most popular frame-based algorithm is the Lucas-Kanade algorithm \cite{Lucas1981}. This algorithm is originally developed for estimating optical flow in the local neighborhood of a pixel. In order to solve \eqref{eq:optical_flow_constraint}, the second assumption is that $u$ and $v$ are constant across neighboring pixels. Therefore, a least-squares system can be composed based on \eqref{eq:optical_flow_constraint}, using $I_x$, $I_y$, and $I_t$ from neighboring pixels. This system can be solved for $u$ and $v$. This technique is mainly applied to sparse estimation, where motion is only computed locally at visual features of interest.

Local optical flow estimation techniques are subjected to the aperture problem, which occurs when motion ambiguity is present due to a limited field of view \cite{Beauchemin1995}. This occurs along object contours which lack clearly distinguishable corner points. The result is that only \emph{normal flow} can be estimated, which is the motion component normal to the contour's orientation. At corner locations, this ambiguity is not present. Only at these points, optical flow is estimated using Lucas-Kanade. Therefore, a corner detection algorithm (e.g. \cite{Rosten2008}) can be applied to first obtain points of interest in the frame. This strategy is applied in many recent optical flow based landing experiments \citet{DeCroon2013, Alkowatly2015, Ho2016, DeCroon2016}.

Alternatively, in \citet{Herisse2012} a 'pyramidal' variant of Lucas-Kanade is applied \cite{Bouguet2000} to account for large displacements. This is a coarse-to-fine approach: optical flow is first computed for a highly downsampled frame. Then, the frame is iteratively refined, computing more detailed optical flow at each refinement level, using the estimate at the previous level to initialize the estimate.

In all these approaches, it is necessary to process full frames, either to find features of interest such as corners, or to obtain sufficiently detailed dense optical flow. While it is possible to use low resolution frames for faster processing, this comes at the cost of reduced detail and hence lower accuracy. Event-based cameras are much less subject to this trade-off, since their output directly highlights locations of interest for estimating optical flow.

\subsubsection{Optical Flow Sensors}
Hardware-based solutions for estimating optical flow have also been applied. Some researchers employ off-the-shelf optical mouse sensors for measuring translational optical flow e.g. \cite{Zufferey2010}. In addition, the visual motion processing in insects inspired researchers to develop highly simplified optical flow sensors, such as the 2-photodetector elementary motion detector was developed for the tethered MAV research in \cite{Ruffier2005,Ruffier2014}. 
Optical flow sensors achieve relatively high sampling rates due to their simplicity. However, their operating principle is generally limited to measuring translational flow. For measuring patterns of optical flow, such as divergence, multiple separate sensors need to be applied and integrated.

\subsubsection{Event-Based Methods}
Since the introduction of the DVS and subsequently developed sensors, several different approaches to event-based optical flow estimation have been developed. Most of these techniques operate on each newly detected event and its spatiotemporal neighborhood, providing sparse optical flow estimates. However, in most cases, the algorithms do not distinguish between corner points and other visual features. Thus, they primarily estimate normal flow. In the following, a brief review of recent approaches is presented.

An adaptation of the frame-based Lucas-Kanade tracker is introduced in \citet{Benosman2012}. Similar to the original algorithm, it solves the optical flow constraint by including the local neighborhood of a pixel. Since absolute measurements of $I$ are not available, the authors replaced the intensity $I$ by the sum of event polarities at a pixel location, obtained over a fixed time window. The reconstructed 'relative intensity' is used to numerically estimate $I_x$, $I_y$, and $I_t$. However, the number of events is generally too low for this approach to provide accurate gradient estimates, in particular for the temporal gradient $I_t$. 

In \citet{Benosman2014} an algorithm is presented that operates on the spatiotemporal representation of events as a point cloud (as shown in \cref{fig:events_frames}). When representing a sequence of events by three-dimensional points of $(x,y,t)$, they form surface-like structures. The gradient of such a surface relates to the motion of the object that triggered the events. By computing a local tangent plane to an event and its neighbor events, normal flow for that event is estimated. A follow-up study employs this algorithm for detecting and tracking corners from neighboring normal flow vectors, hence obtaining fully observable optical flow \cite{Clady2015}. However, real-time results are not yet demonstrated with a non-parallelized implementation.
%
%

In \cite{Barranco2014} a technique is introduced that estimates optical flow on object contours, based on both events and absolute intensity measurements. Input events are used to locate motion boundaries on contours. Along each boundary, motion is estimated using the width of the contour, which is computed from the local event distribution and the absolute intensity. The latter can be reconstructed from events, but having separate intensity measurements (e.g. from a DAVIS or ATIS sensor) simplifies the process.

A bio-inspired approach is proposed in \citet{Brosch2015}. In this approach, optical flow is estimated using direction- and speed-selective filters based on the first stages of visual processing in humans. A bank of spatiotemporal filters is employed, each of which is maximally selective for a certain direction and speed of optical flow. For each new event, the neighboring event cloud is convolved with the filters to obtain a confidence measure for each filter. Optical flow for that event is then obtained from the sum of the confidence measures weighted by direction.

More complex event-based algorithms have also been developed, which have not demonstrated real-time performance, but show promising results. In \citet{Barranco2015} a phase-based optical flow method is discussed, which is developed for high-frequency textures. The algorithm is compared to other event-based methods \cite{Benosman2012,Benosman2014,Barranco2014}, indeed showing significant accuracy improvements. Also, an approach was presented for simultaneous estimation of dense optical flow and absolute intensity \cite{Bardow2016}. This is the only available approach aimed towards dense optical flow estimation. Visual results of this method are encouraging, yet a quantitative evaluation is not performed.

Recently, several datasets for event-based visual navigation have been published. The set in \citet{Barranco2016} provides both frame and event measurements from a DAVIS sensor accompanied by odometry measurements. This facilitates comparison between frame-based and event-based techniques for optical flow estimation or visual odometry. However, to the best of the authors' knowledge, an actual comparison of existing techniques has not yet been published for this set. In this respect, the work in \citet{Ruckauer2016} is more relevant for this work, as it features both an event-based dataset and a comparison of various optical flow algorithms. These are variants of the techniques in \cite{Benosman2012} and \cite{Benosman2014}, as well as a basic direction selective algorithm.

We select the local plane fitting algorithm in \citet{Benosman2014} as the basis of the approach in our work. It has shown the most promising results in \cite{Ruckauer2016} and has recently been incorporated into follow-up experiments \cite{Clady2014,Clady2015}. In addition, its implementations yielded real-time operation for high event measurement rates.



\section{Relations between Optical Flow, Ego-Motion, and Visual Observables}
\label{sec:model}

This section defines the optical flow model, which relates the ego-motion of an MAV equipped with a downward facing camera to the perceived optical flow, and the visual observables presented in \cref{sec:related_landing}. These relations form the basis for our evaluation methods applied in subsequent sections, in which ground truth values for optical flow and visual observables are computed.

\begin{figure}[!t]
	\centering
	\includegraphics[width=0.5\linewidth]{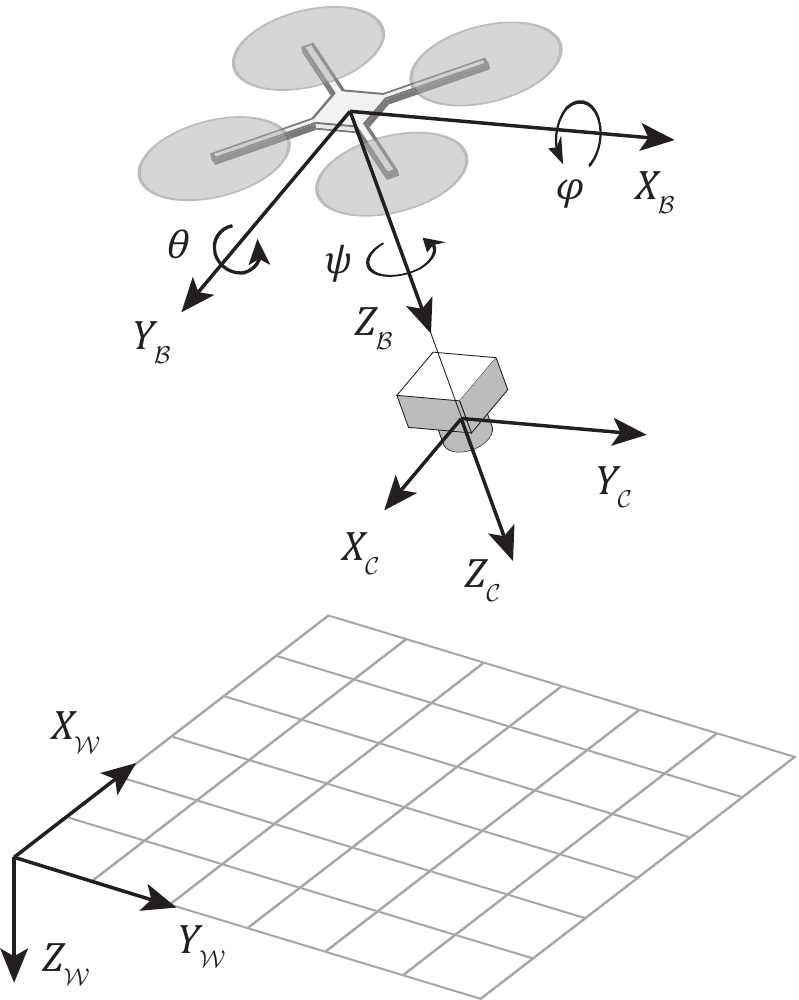}
	\caption{Definitions of the world ($\cal W$), body ($\cal B$), and camera ($\cal C$) reference frames. The Euler angle definitions and their signs are also shown.}
	\label{fig:referenceframes}
\end{figure}

In the derivation, use is made of three reference frames: $\mathcal{B}$, $\mathcal{C}$, and $\mathcal{W}$, which describe the body, camera and inertial world reference frames respectively. Their definitions are illustrated in \cref{fig:referenceframes}. In each reference frame, a position is denoted through the coordinates $\left (X,Y,Z\right)$, with corresponding velocity components $\left (U,V,W\right)$. The body frame is centered at the center of gravity of the MAV. The rotation from $\mathcal{W}$ to $\mathcal{B}$ is described by standard Euler angles $\varphi$, $\theta$, $\psi$, denoting roll, pitch, and yaw respectively. Similarly, $p$, $q$, and $r$ describe the roll, pitch, and yaw rotational rates.

The camera reference frame $\cal C$ is centered at the focal point of the DVS. The camera is assumed to be located directly below the MAV's center of gravity, with $X_\mathcal{C} = Y_\mathcal{B}$ and $Y_\mathcal{C} = -X_\mathcal{B}$. However, we account for an offset $\Delta Z$ between $Z_\mathcal{C}$ and $Z_\mathcal{B}$, i.e. $Z_\mathcal{C} = Z_\mathcal{B}+\Delta Z$.

The relations between optical flow and ego-motion are based on the pinhole camera model formulation in \citet{Longuet-Higgins1980}. In this formulation, pixel locations $\left(x,y\right)$ in the sensor's pixel grid and optical flow components $\left(u,v\right)$ in pixels per second are represented by their metric, real-world equivalents $\left(\hat x, \hat y\right)$ and $\left(\hat u, \hat v\right)$. The mapping between the two representations is composed of two parts: correction for lens distortion and transformation through the camera's intrinsics matrix. A two-parameter version of the commonly used Brown-Conrady model \cite{Brown1966} is applied to model the relation between $\left(x,y\right)$ and their undistorted equivalent $\left(x_u,y_u\right)$: 

\begin{equation}
\label{eq:distortion}
{\left[ {\begin{array}{*{20}{c}}
		{{x-x_p}}\\{{y-y_p}}
		\end{array}} \right]} = {\left[ {\begin{array}{*{20}{c}}
		x_u-x_p\\y_u-y_p
		\end{array}} \right]}\left( {1 + {k_1}{r_u^2} + {k_2}{r_u^4}} \right)
\end{equation}

The point $\left(x_p,y_p\right)$ in \cref{eq:distortion} represents the camera's principal point, and $r_u=\sqrt{\left(x_u-x_p\right)^2+\left(y_u-y_p\right)^2}$ is the radial distance to the principal point. Second, the undistorted pixels and corresponding optical flow estimates relate to their metric equivalent through scaling with the focal length $f$:
\begin{equation}
\hat x = \frac{{x_u - {x_p}}}{f},\quad \hat y = \frac{{y_u - {y_p}}}{f},\quad \hat u = \frac{u}{f},\quad \hat v = \frac{v}{f}
\end{equation}
In order to obtain the parameters $k_1$, $k_2$, $x_p$, $y_p$, and $f$, the DVS is calibrated using the Camera Calibration Toolbox in MATLAB \cite{Bouguet1999}. Since the DVS does not record absolute intensity, artificial images are generated by recording events from a flashing checkerboard pattern on an LCD screen, similar to \citet{Mueggler2014}.

The optical flow components $\left(\hat{u},\hat{v}\right)$ due to an arbitrary point moving in $\cal C$ with motion $\left(U_{\cal C},V_{\cal C}, W_{\cal C}\right)$ at depth $Z_{\cal C}$ are obtained as follows:

\begin{equation}
\label{eq:optical_flow_equations}
\begin{aligned}
\hat u &=  - \frac{{{U_{\cal C}}}}{{{Z_{\cal C}}}} + \hat x\frac{{{W_{\cal C}}}}{{{Z_{\cal C}}}} - p + r\hat y + q\hat x\hat y - p{{\hat x}^2}\\
\hat v &=  - \frac{{{V_{\cal C}}}}{{{Z_{\cal C}}}} + \hat y\frac{{{W_{\cal C}}}}{{{Z_{\cal C}}}} + q - r\hat x + q{{\hat y}^2} - p\hat x\hat y
\end{aligned}
\end{equation}

When separate measurements of the rotational rates $p$, $q$, and $r$ are available (for instance, from rate gyro measurements), the optical flow in \cref{eq:optical_flow_equations} can be corrected for the ego-rotation of the camera. This derotation leaves only the translational optical flows $\hat u_T$, $\hat v_T$. Further, if all visible points are part of a single plane, their coordinates $Z_\mathcal{C}$ are interrelated. In most indoor applications, floor surfaces can be assumed planar and horizontal. However, MAVs perform fast horizontal maneuvers through rolling and pitching, such that the ground plane may have a slight inclination in $\mathcal{C}$. In this case, $Z_\mathcal{C}$ is expressed through three parameters: the distance $Z_0$ to the plane at $\left(\hat{x},\hat{y}\right)=\left(0,0\right)$, and the plane slopes ${Z_X}$, ${Z_Y}$. 

\begin{equation}
\label{eq:plane}
Z_\mathcal{C} = Z_0 + Z_X X_\mathcal{C} + Z_Y Y_\mathcal{C}
\end{equation}

This can be rewritten into:

\begin{equation}
\label{eq:plane2}
\frac{Z_\mathcal{C}-Z_0}{Z_\mathcal{C}}=Z_X\hat{x}+Z_Y\hat{y}
\end{equation}

When the surface is flat, the slopes are the tangents of the roll and pitch angles, and can therefore be obtained from separate sensors in e.g. an IMU:

\begin{equation}
{Z_X} =  - \tan \varphi ,\quad {Z_Y} = \tan \theta
\end{equation}

Now we define the \emph{scaled velocities} $\vartheta_x$, $\vartheta_y$, $\vartheta_z$ as follows:

\begin{equation}
\label{eq:normalized_velocities}
{\vartheta _x} =  \frac{U_\mathcal{C}}{Z_0},\quad {\vartheta _y} =  \frac{V_\mathcal{C}}{Z_0},\quad {\vartheta _z} =  \frac{W_\mathcal{C}}{Z_0}
\end{equation}

Following the derivation in \citet{DeCroon2013}, substituting \cref{eq:plane2} and \cref{eq:normalized_velocities} into \cref{eq:optical_flow_equations} leads to the following expression:

\begin{equation}
\label{eq:planar_flow_field1}
\begin{aligned}
\hat u_T &= \left( { - {\vartheta _x} + \hat x{\vartheta _z}} \right)\left( {1-Z_X\hat{x}-Z_Y\hat{y}} \right) \\
\hat v_T &= \left( { - {\vartheta _y} + \hat y{\vartheta _z}} \right)\left( {1-Z_X\hat{x}-Z_Y\hat{y}} \right)
\end{aligned}
\end{equation}

The scaled velocities represent the visual observables presented in \cref{sec:related_landing}. $\vartheta_x$ and $\vartheta_y$ are the opposites of the ventral flows, i.e. $\omega_x = -\vartheta_x$, $\omega_y = -\vartheta_y$. The vertical component $\vartheta_z$ is proportional to the flow field divergence, since $D=\nabla\cdot\mathbf{V}=2\vartheta_z$. It is also the inverse of time-to-contact $\tau = Z_{\cal C}/W_{\cal C} = 1/\vartheta_z$.

Note that, in the presence of the vertical camera offset $\Delta Z$, the rotational rates $p$ and $q$ induce additional translational velocity components into $U_{\cal C}$ and $V_{\cal C}$. These propagate to $\vartheta_x$ and $\vartheta_y$, such that:

\begin{equation}
\vartheta_x = \frac{V_{\cal B}}{Z_{\cal C}} - p\frac{\Delta Z}{Z_{\cal C}},\quad \vartheta_y = \frac{U_{\cal B}}{Z_{\cal C}} + q\frac{\Delta Z}{Z_{\cal C}}
\end{equation}

These corrections are accounted for in the computation of ground truth optical flow and values of $\vartheta_x$ and $\vartheta_y$.

\section{Event-Based Optical Flow Estimation}
\label{sec:eof_estimation}

This section describes our optical flow estimation approach. Since it is based on the work in \citet{Benosman2014}, this baseline approach is explained first in \cref{sec:eof_plane_fitting}. Then, the proposed modifications for achieving higher efficiency (\cref{sec:eof_efficiency_improvements}) and timestamp-based selection (\cref{sec:eof_timesorting}) are discussed. In \cref{sec:results_optical_flow} the result of our improvements is evaluated in comparison to the baseline algorithm.

\subsection{The Baseline Plane Fitting Algorithm}
\label{sec:eof_plane_fitting}
The main working principle of the baseline algorithm is based on the space-time representation of events as a point cloud. In the following, an event is denoted as a space-time point according to $\mathbf{e}_n=\left(x,y,t\right)$, where $x$ and $y$ represent the undistorted pixel locations. Note that the polarity $P$ is not considered; we group positive and negative polarity events and process them separately.

Let $\Sigma_e (x,y) = t$ be a mapping describing the surface along which events are positioned. The shape of $\Sigma_e$ is a result of the feature geometry and, in particular, its motion. In the case of a locally linear feature (such as an edge) and constant motion, this surface reduces to a plane. This is clearly visible in the example scene in \cref{fig:events_frames}. With these assumptions, $\Sigma_e$ can be approximated by a tangent plane within a limited range of $x$, $y$, and $t$.

For each newly detected event $\mathbf{e}_n$, a plane $\mathbf{\Pi}$ is computed that fits best to all neighboring events $\mathbf{e}_i$ for which $x_i \in \left[x_n-\frac{1}{2}\Delta x,\, x_n+\frac{1}{2}\Delta x\right]$, $y_i \in \left[y_n-\frac{1}{2}\Delta y,\, y_n+\frac{1}{2}\Delta y\right]$, and $t_i \in \left[t_n-\Delta t,\, t_n\right]$, where $\Delta x, \Delta y, \Delta t$ indicate spatial and temporal windows. The spatial windows are generally small and are both set to 5 pixels. The temporal window setting has a large influence on the detectable speed and interference of multiple features, which is discussed further in \cref{sec:eof_timesorting}. 

The plane $\mathbf{\Pi} = \left[p_x,\, p_y,\, p_t,\, p_0\right]^T$ is computed through an iterative process of linear least squares regression and outlier rejection. It is represented in homogeneous coordinates, such that the following hold for any event $\mathbf{e}_i$ that intersects with $\mathbf{\Pi}$:

\begin{equation}
\label{eq:plane_system}
p_x x_i + p_y y_i + p_t t_i + p_0 = 0
\end{equation}

Extending \cref{eq:plane_system} with at least four neighboring events, an overdetermined system of equations is obtained, which is solved through linear least-squares. After an initial fit, the Euclidean distance of each event to the plane is computed. All events for which the distance exceeds a threshold $d_{max}$, are rejected from this fit. Using the remaining events, a new least-squares plane fit is computed. In \citet{Benosman2014}, this process is repeated until the change in $\mathbf{\Pi}$ is no longer significant. This is the case if the norm of the change in all components in $\mathbf{\Pi}$ is smaller than a second threshold $k_d$, i.e. $\left\| {\mathbf{\Pi} (i) - \mathbf{\Pi} (i - 1)} \right\| < k_d$. In practice, the latter often occurs already after one or two iterations. In this work, the values for $d_{max}$ and $k_d$ specified in \citet{Ruckauer2016} are applied, which are 0.01 and 1e-5 respectively.

The final plane is preserved and used to compute the local gradients of $\Sigma_e$:

\begin{equation}
\nabla {\Sigma _e}\left( {x,y} \right) = {\left[ {\frac{{\partial {\Sigma _e}}}{{\partial x}},\,\frac{{\partial {\Sigma _e}}}{{\partial y}}} \right]^T} = \left[-\frac{p_x}{p_t},\, -\frac{p_y}{p_t} \right]^T
\end{equation}

In \citet{Benosman2014} the gradient components of $\Sigma_e$ are assumed to be inversely related to the optical flow components $(u,v)$:
\begin{equation}
\label{eq:eof_slopes_1}
\nabla {\Sigma _e}\left( {x,y} \right)  = {\left[ {\frac{1}{{u\left( {x,y} \right)}},\,\frac{1}{{v\left( {x,y} \right)}}} \right]^T}
\end{equation}

However, as is also noted in \citet{Brosch2015,Ruckauer2016}, \cref{eq:eof_slopes_1} is subject to singularities when computing $u$ and $v$. If either component of $\nabla\Sigma_e$ tends to zero, the corresponding optical flow component grows to infinity, which is incorrect. For example, consider a horizontally moving vertical line. Along the $y$-direction, temporal differences between the resulting events are in this case very small. Therefore, $\frac{{\partial {\Sigma _e}}}{{\partial y}}$ is also small, which leads to a high value of the vertical component $v$, even though the line is moving horizontally.

In recent work, two modifications to the previously discussed methodology have been proposed. First, in \citet{Brosch2015} an approach is presented that is robust to singularities in $u$ and $v$, which led to significant accuracy improvements in the comparison in \citet{Ruckauer2016}. In this approach, an orthogonality constraint is imposed on the plane's normal vector $\left[p_x,\,p_y,\,p_t\right]^t$, the optical flow vector $\left[u,\,v,\,1\right]$ and the orientation $\left[l_x,\,l_y,\,0\right]$ of the edge in homogeneous coordinates. This constraint leads to a new expression of the optical flow components $u$ and $v$ in terms of the plane's normal vector:

\begin{equation}
\label{eq:slopes_to_flow}
\left[ {\begin{array}{*{20}{c}}
	u\\
	v
	\end{array}} \right] = \frac{1}{{{{\left\| {\nabla {\Sigma _e}} \right\|}^2}}}\nabla {\Sigma _e} = -\frac{{{p_t}}}{{{p_x}^2 + {p_y}^2}}\left[ {\begin{array}{*{20}{c}}
	{{p_x}}\\
	{{p_y}}
	\end{array}} \right]
\end{equation}

Second, in the implementation in \cite{Ruckauer2016} not all events within the space-time window are considered. For each pixel location, only the most recent event is used for computing optical flow. However, high contrast edges tend to produce multiple events in quick succession at a single pixel. Hence, the most recent event at a pixel occurs slightly later than the first event caused by such an edge, which leads to over-estimation of its speed. It may also lead to optical flow estimates in the opposite direction of the edge motion. To prevent this, a refractory period $\Delta t_R$ (typically 0.1 s) is applied. Events that occur within $\Delta t_R$ are neither processed nor preserved to support future events.

The discussed algorithm with the previously proposed modifications forms our baseline algorithm. In the following, methods are proposed to increase its efficiency and range of application. 

\subsection{Efficiency Improvements}
\label{sec:eof_efficiency_improvements}
In order to enable faster computation and scale the algorithm towards low-end processing hardware, we propose two modifications. 

The first modification is to reduce the number of parameters of the local plane. We reduce \cref{eq:plane_system} by introducing the new parameters $p_x^\ast$, $p_y^\ast$, $p_0^\ast$:

\begin{equation}
p_x^ *  = \frac{{{p_x}}}{{{p_t}}},\, p_y^ *  = \frac{{{p_y}}}{{{p_t}}},\, p_0^ *  = \frac{{{p_0}}}{{{p_t}}}
\end{equation}

hence obtaining the nonhomogeneous, three-parameter form of \cref{eq:plane_system}:

\begin{equation}
\label{eq:plane_system_2}
p_x^* x_i + p_y^* y_i + p_0^* = -t_i
\end{equation}

A second reduction is performed by assuming that the new event $\mathbf{e}_n$ intersects with the plane, which enables the definition of relative coordinates for neighbor events as follows. Given $\mathbf{e}_n$ and a previously identified neighbor event $\mathbf{e}_i$, the relative coordinates of the neighbor event are defined as $\delta x_i = x_i - x_n,\; \delta y_i = y_i - y_n,\; \delta t_i = t_i - t_n$. Substituting these coordinates into \cref{eq:plane_system_2} and rearranging terms, the following relation is obtained: 

\begin{equation}
\label{eq:plane_system_22}
p_x^*\delta {x_i} + p_y^*\delta {y_i} + \delta {t_i} =  - p_x^*{x_n} - p_y^*{y_n} - p_0^* - {t_n}
\end{equation}

By substituting \cref{eq:plane_system_2} (for which we set $i=n$ to enforce that $\emph{e}_n$ intersects with the plane) into \cref{eq:plane_system_22}, the right-hand side of the latter equation reduces to zero. Thus, the final plane $\mathbf{\Pi}^\ast=\left[p_x^\ast,\, p_y^\ast\right]$ is described by two parameters, the slopes:

\begin{equation}
\label{eq:plane_system_3}
p_x^* \delta x_i + p_y^* \delta y_i =  - \delta t_i
\end{equation}

This reduced approach requires significantly less computational effort than the baseline. While solving a homogeneous least squares system is generally performed using a Singular Value Decomposition (SVD), more efficient solvers such as the commonly used QR-decomposition are applicable to nonhomogeneous problems. With a system of $M$ events and $N$ parameters, the computational complexity of the SVD scales with $O(MN^2+N^3)$. In comparison, the complexity of the QR decomposition scales with $O(MN^2-N^3/3)$ \cite{Heath2002}. Hence, a four-parameter SVD solution has a cost that is approximately proportional to $16M+64$, which compares to $4M-8/3$ for a two-parameter QR-decomposition. Note that this is only a rough indication of the true complexity, but it suffices for illustrating the efficiency gain of the parameter reduction. Only a slight reduction in accuracy is introduced with this simplification.

The second modification consists of capping the rate at which optical flow vectors are identified, denoted as the output rate $\rho_F$. Depending on the computational resources available, input events can be processed at a limited rate to maintain real-time performance. In addition, since the approach assumes that for each individual event, motion needs to be estimated, neighboring events produce highly similar optical flow vectors, making the information increasingly redundant with increasing $\rho_F$. Therefore, capping this value also prevents unnecessary computational load on follow-up processes. To achieve this, we keep track of the timestamp $t_f$ of the event for which the latest optical flow vector was identified. If a new event $\mathbf{e}_n$ occurs, optical flow is only estimated if $t_n - t_f > 1/\rho_{F_{max}}$, where $\rho_{F_{max}}$ denotes the output rate limit. The event is, however, still stored to support future events, taking into account the refractory period. Hence, accuracy of the estimates that are still performed, is unaffected. The resulting effect of the value of $\rho_{F_{max}}$ on computational performance is explored in \cref{sec:results_optical_flow_computation}.

\subsection{Timestamp-Based Clustering of Recent Events}
\label{sec:eof_timesorting}

%

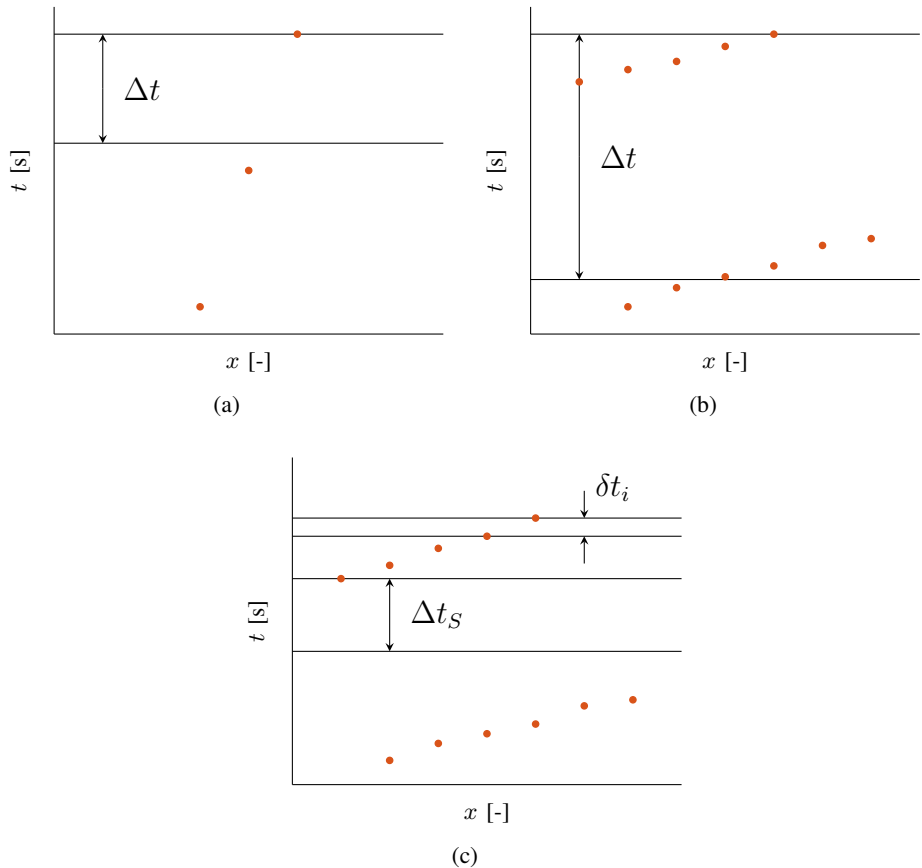
\begin{figure}[!t]
	\centering
	\setlength{\fwidth}{0.3\linewidth}
	\renewcommand{\ylabeldist}{0.15}
	\subfloat[]{
%
%
\definecolor{mycolor1}{rgb}{0.85000,0.32500,0.09800}%
\begin{tikzpicture}[%
arrow1/.style={->,color=black,solid}
]

\begin{axis}[%
width=0.951\fwidth,
height=0.8\fwidth,
at={(0\fwidth,0\fwidth)},
scale only axis,
xmin=0.0000,
xmax=4.0000,
xtick style={draw=none},
xtick={0.0000,1.3333,2.6667,4.0000},
xticklabels={\empty},
xlabel={$x$ {[-]}},
ymin=-0.2000,
ymax=2.2000,
ytick style={draw=none},
ytick={-0.2000,0.6000,1.4000,2.2000},
yticklabels={\empty},
ylabel={$t$ {[s]}},
axis background/.style={fill=white},
axis x line*=bottom,
axis y line*=left,
title style={font=\labelsize},
xlabel style={font=\labelsize,at={(axis description cs:0.5,\xlabeldist)}},
ylabel style={font=\labelsize,at={(axis description cs:\ylabeldist,0.5)}},
legend style={font=\ticksize},
ticklabel style={font=\ticksize}
]
\addplot[only marks,mark=*,mark options={line width=0.5pt},mark size=1.1785pt,color=mycolor1] plot table[row sep=crcr]{%
1.5000	0.0000\\
2.0000	1.0000\\
2.5000	2.0000\\
};
\addplot [color=black,solid,forget plot]
  table[row sep=crcr]{%
0.0000	2.0000\\
4.0000	2.0000\\
};
\addplot [color=black,solid,forget plot]
  table[row sep=crcr]{%
0.0000	1.2000\\
4.0000	1.2000\\
};
\addplot [arrow1] coordinates{(0.5000,1.6000) (0.5000,1.2000)};
\addplot [arrow1] coordinates{(0.5000,1.6000) (0.5000,2.0000)};
\node[right, align=left, text=black]
at (axis cs:0.6,1.6) {$\Delta t$};
\end{axis}
\end{tikzpicture}%
		\label{fig:time_window_1}
	}
	\subfloat[]{
%
%
\definecolor{mycolor1}{rgb}{0.85000,0.32500,0.09800}%
\begin{tikzpicture}[%
arrow1/.style={->,color=black,solid}
]

\begin{axis}[%
width=0.951\fwidth,
height=0.8\fwidth,
at={(0\fwidth,0\fwidth)},
scale only axis,
xmin=0.0000,
xmax=4.0000,
xtick style={draw=none},
xtick={0.0000,1.3333,2.6667,4.0000},
xticklabels={\empty},
xlabel={$x$ {[-]}},
ymin=-0.2000,
ymax=2.2000,
ytick style={draw=none},
ytick={-0.2000,0.6000,1.4000,2.2000},
yticklabels={\empty},
ylabel={$t$ {[s]}},
axis background/.style={fill=white},
axis x line*=bottom,
axis y line*=left,
title style={font=\labelsize},
xlabel style={font=\labelsize,at={(axis description cs:0.5,\xlabeldist)}},
ylabel style={font=\labelsize,at={(axis description cs:\ylabeldist,0.5)}},
legend style={font=\ticksize},
ticklabel style={font=\ticksize}
]
\addplot[only marks,mark=*,mark options={line width=0.5pt},mark size=1.1785pt,color=mycolor1] plot table[row sep=crcr]{%
0.5000	1.6500\\
1.0000	1.7400\\
1.5000	1.8000\\
2.0000	1.9100\\
2.5000	2.0000\\
1.0000	0.0000\\
1.5000	0.1400\\
2.0000	0.2200\\
2.5000	0.3000\\
3.0000	0.4500\\
3.5000	0.5000\\
};
\addplot [color=black,solid,forget plot]
  table[row sep=crcr]{%
0.0000	2.0000\\
4.0000	2.0000\\
};
\addplot [color=black,solid,forget plot]
  table[row sep=crcr]{%
0.0000	0.2000\\
4.0000	0.2000\\
};
\addplot [arrow1] coordinates{(0.5000,1.1000) (0.5000,0.2000)};
\addplot [arrow1] coordinates{(0.5000,1.1000) (0.5000,2.0000)};
\node[right, align=left, text=black]
at (axis cs:0.6,1.1) {$\Delta t$};
\end{axis}
\end{tikzpicture}%
		\label{fig:time_window_2}
	}\\
	\subfloat[]{
%
%
\definecolor{mycolor1}{rgb}{0.85000,0.32500,0.09800}%
\begin{tikzpicture}[%
arrow1/.style={->,color=black,solid},
arrow2/.style={->,color=black,solid}
]

\begin{axis}[%
width=0.951\fwidth,
height=0.8\fwidth,
at={(0\fwidth,0\fwidth)},
scale only axis,
xmin=0.0000,
xmax=4.0000,
xtick style={draw=none},
xtick={0.0000,1.3333,2.6667,4.0000},
xticklabels={\empty},
xlabel={$x$ {[-]}},
ymin=-0.2000,
ymax=2.5000,
ytick style={draw=none},
ytick={-0.2000,0.7000,1.6000,2.5000},
yticklabels={\empty},
ylabel={$t$ {[s]}},
axis background/.style={fill=white},
axis x line*=bottom,
axis y line*=left,
title style={font=\labelsize},
xlabel style={font=\labelsize,at={(axis description cs:0.5,\xlabeldist)}},
ylabel style={font=\labelsize,at={(axis description cs:\ylabeldist,0.5)}},
legend style={font=\ticksize},
ticklabel style={font=\ticksize}
]
\addplot[only marks,mark=*,mark options={line width=0.5pt},mark size=1.1785pt,color=mycolor1] plot table[row sep=crcr]{%
0.5000	1.5000\\
1.0000	1.6100\\
1.5000	1.7500\\
2.0000	1.8500\\
2.5000	2.0000\\
1.0000	0.0000\\
1.5000	0.1400\\
2.0000	0.2200\\
2.5000	0.3000\\
3.0000	0.4500\\
3.5000	0.5000\\
};
\addplot [color=black,solid,forget plot]
  table[row sep=crcr]{%
0.0000	2.0000\\
4.0000	2.0000\\
};
\addplot [color=black,solid,forget plot]
  table[row sep=crcr]{%
0.0000	1.8500\\
4.0000	1.8500\\
};
\addplot [arrow1] coordinates{(3.0000,2.2250) (3.0000,2.0000)};
\addplot [arrow1] coordinates{(3.0000,1.6250) (3.0000,1.8500)};
\node[right, align=left, text=black]
at (axis cs:3,2.25) {$\delta t_i$};
\addplot [color=black,solid,forget plot]
  table[row sep=crcr]{%
0.0000	1.5000\\
4.0000	1.5000\\
};
\addplot [color=black,solid,forget plot]
  table[row sep=crcr]{%
0.0000	0.9000\\
4.0000	0.9000\\
};
\addplot [arrow2] coordinates{(1.0000,1.2000) (1.0000,0.9000)};
\addplot [arrow2] coordinates{(1.0000,1.2000) (1.0000,1.5000)};
\node[right, align=left, text=black]
at (axis cs:1.1,1.2) {$\Delta t_S$};
\end{axis}
\end{tikzpicture}%
		\label{fig:time_window_3}
	}
	\caption{Examples with one-dimensional ($x-t$) event structures representing motion, in which a fixed time window for collecting events leads to failed motion estimates. In \protect\subref{fig:time_window_1}, the time window is too small for being able to perceive the slow motion that triggers the events. On the other hand, in \protect\subref{fig:time_window_2} events from two sequential fast-moving features enter the same time window. The bottom image \protect\subref{fig:time_window_3} illustrates the proposed clustering approach, in which the time difference between the current event and the next most recent event $\delta t_i$ defines the maximum time difference $\Delta t_S=k_S \delta t_i$. From the leftmost event, the time difference to the next most recent event exceeds $\Delta t_S$, such that all bottom events are rejected.}
	\label{fig:time_window}
\end{figure}

The baseline algorithm incorporates a fixed setting for the time window $\Delta t$ to collect recent events. There are two main drawbacks of using a fixed time window, which are illustrated in \cref{fig:time_window} for two simple one-dimensional cases. First, $\Delta t$ defines the lower limit for the magnitude of observable optical flow. For slower motion, the time difference between neighboring events increases. If this difference is too large, all neighboring events fall outside the time window (as illustrated in \cref{fig:time_window_1}), such that the motion cannot be observed. Second, a larger time window can result in the inclusion of unrelated events. For example, in \cref{fig:time_window_2} events are shown which clearly belong to separate features. Still, a part of the outdated features falls within the time window, which leads to an inaccurate fit. In some cases outlier rejection may prevent this, but with tightly packed features this may still cause a failed estimate. A fixed time window therefore imposes a fundamental trade-off between minimal observable speed and feature density. Since MAVs tend to move at a wide range of velocities, from hovering to fast maneuvers, the capability of observing both fast and slow motion is desirable.

%

To accomplish this, we propose a very simple clustering method based on the time order of events, which is illustrated in \cref{fig:time_window_3} for a one-dimensional motion case. First, the minimum number of most recent events $\mathbf{e}_i$ for observing velocity is found. In the one-dimensional case in \cref{fig:time_window_3}, only one point is necessary for constructing a line; for two-dimensional image motion, two linearly independent points $(\delta x_i,\delta y_i,\delta t_i)$ are required in order to construct a plane. From the point with the largest $\delta t_i$, the relative timestamp defines a maximum time increment $\Delta t_S$ between the timestamps of consecutive events. To provide a margin for noise, $\delta t_i$ is scaled with a factor $k_S$ (which has a value of 3 in our experiments), such that $\Delta t_S=-\delta t_i k_S$ (since $\delta t_i$ should be negative). Second, we iterate through the remaining recent events, ordered by decreasing value of $\delta t_i$. If the time difference between two consecutive events $\mathbf{e}_i$ and $\mathbf{e}_{i-1}$ exceeds $\Delta t_S$, $\mathbf{e}_{i-1}$ and all events that occurred before it are assumed to belong to different features, and are rejected. 

Note that this approach does not take spatial location into account, except for finding the first linearly independent events. Therefore, a variation on the baseline process of outlier rejection is still applied, which is independent of time-scale. Instead of rejection based on a distance threshold, the overall fit quality is assessed through the Normalized Root Mean Square Error ($\mathit{NMRSE}$), defined here as follows:

\begin{equation}
\label{eq:NRMSE}
\mathit{NRMSE} = \frac{n}{{\sum\limits_{i = 1}^n {\delta {t_i}} }}\sqrt {\frac{{\sum\limits_{i = 1}^n {{{\left( {\delta {t_i} - p_x^*\delta {x_i} - p_y^*\delta {y_i}} \right)}^2}} }}{n}} 
\end{equation}

Then, while $\mathit{NRMSE} > \mathit{NRMSE_{max}}$, only the event having the maximum distance to $\mathbf{\Pi}^*$ is rejected. This is repeated until a maximum number of $n_{R}$ events are rejected. If $n_R$ is exceeded, the estimated plane is rejected and no optical flow is computed. Suitable values for attaining a high number of successful estimates, without sacrificing significant quality, are empirically set at $\mathit{NRMSE_{max}}=0.3$ and $n_R=2$. 

Still, incorrect optical flow estimates may be detected, either due to noise in the event stream or due to undesired inclusion of outlier events. Depending on the application, certain optical flow magnitudes can be deemed unrealistic in advance. Optical flow estimates are therefore rejected if their magnitude exceeds a threshold $V_{max}$, which is set to 1000 pixels/s. In addition, a minimum number of events $n_{min}$ must be found through the clustering mechanism in order to have sufficient support for a reliable fit. This number is set to 8 events. Further, a maximal time window of $\Delta t=2$ s is maintained such that unnecessary event checking is prevented. Note, however, that this time window can now be much larger than in the baseline approach. 

%
%

\subsection{Evaluation}
\label{sec:results_optical_flow}
To evaluate optical flow estimation performance, several datasets were recorded in which the DVS was moved by hand, facing towards a ground surface covered with a textured mat. The recordings are performed indoors, using an Optitrack motion tracking system to measure ground truth position and orientation of the DVS. From these measurements, ground truth optical flow vectors are obtained using the relations derived in \cref{sec:model}. This is performed for each event for which optical flow is identified, hence providing the means for quantitative accuracy evaluation. Although currently datasets are already available \cite{Barranco2016,Ruckauer2016}, the recorded set specifically represents motion above a flat surface in indoor lighting conditions, i.e. the environment in which flight tests are performed in \cref{sec:results}.

\begin{figure}[!ht]
	\centering
	\subfloat[Checkerboard]{
		\includegraphics[width=0.3\linewidth]{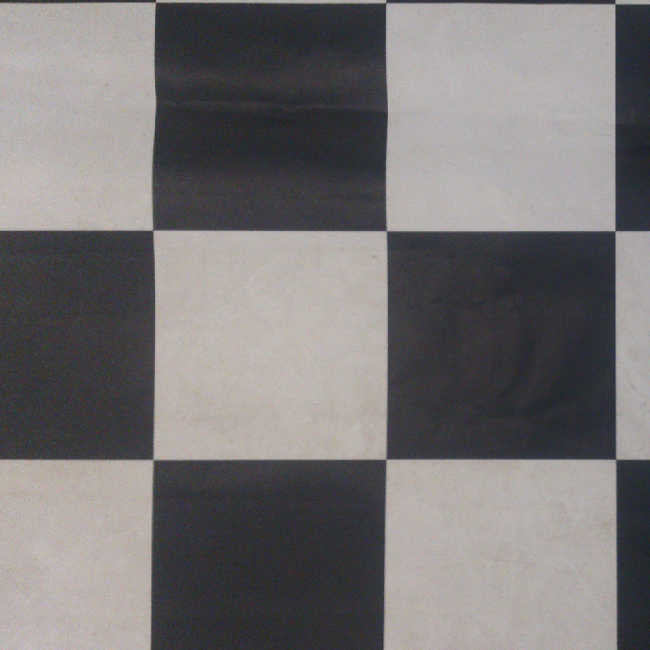}
		\label{fig:checkerboard}
	}
	\subfloat[Roadmap]{
		\includegraphics[width=0.3\linewidth]{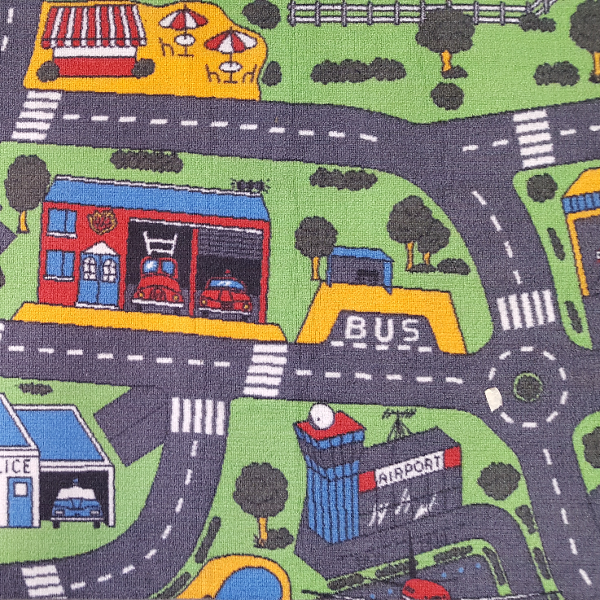}
		\label{fig:roadmap}
	}
	\caption{Ground surface textures used during the experiments.}
	\label{fig:textures}
\end{figure}

Images of the recorded ground surfaces are shown in \cref{fig:textures}. The checkerboard in \cref{fig:checkerboard} provides high contrast and clear edges and is hence relatively simple for optical flow estimation. The roadmap texture in \cref{fig:roadmap} has largely unstructured features and lower contrast. It is used to show that our approach extends to more general situations as well. 
Eight short sequences were selected to evaluate the performance of the proposed method. Each sequence is approximately 1.0 s long and consists of event and pose measurements in which one primary motion type is present. Five sets are selected for the checkerboard; one for vertical translational image motion ($\vartheta_y \approx 1.0$), one for rotational motion ($r\approx-1.3$) rad/s, and three sets with diverging motion of different speeds ($\vartheta_z \approx \lbrace0.2,\,0.5,\,2.0\rbrace$). For the roadmap texture, three sets with diverging motion were selected as well ($\vartheta_z \approx \lbrace0.1,\,0.5,\,1.0\rbrace$). 

In the following analysis, the cap on $\rho_F$ is not applied (i.e. $\rho_{F_{max}}=\infty$), except for the assessment of computational complexity in \cref{sec:results_optical_flow_computation}.

%
%

\subsubsection{Qualitative evaluation}
\begin{figure*}[!ht]
	\begin{framed}
		\centering
		\subfloat[Translating checkerboard]{
			\includegraphics[width=0.31\linewidth]{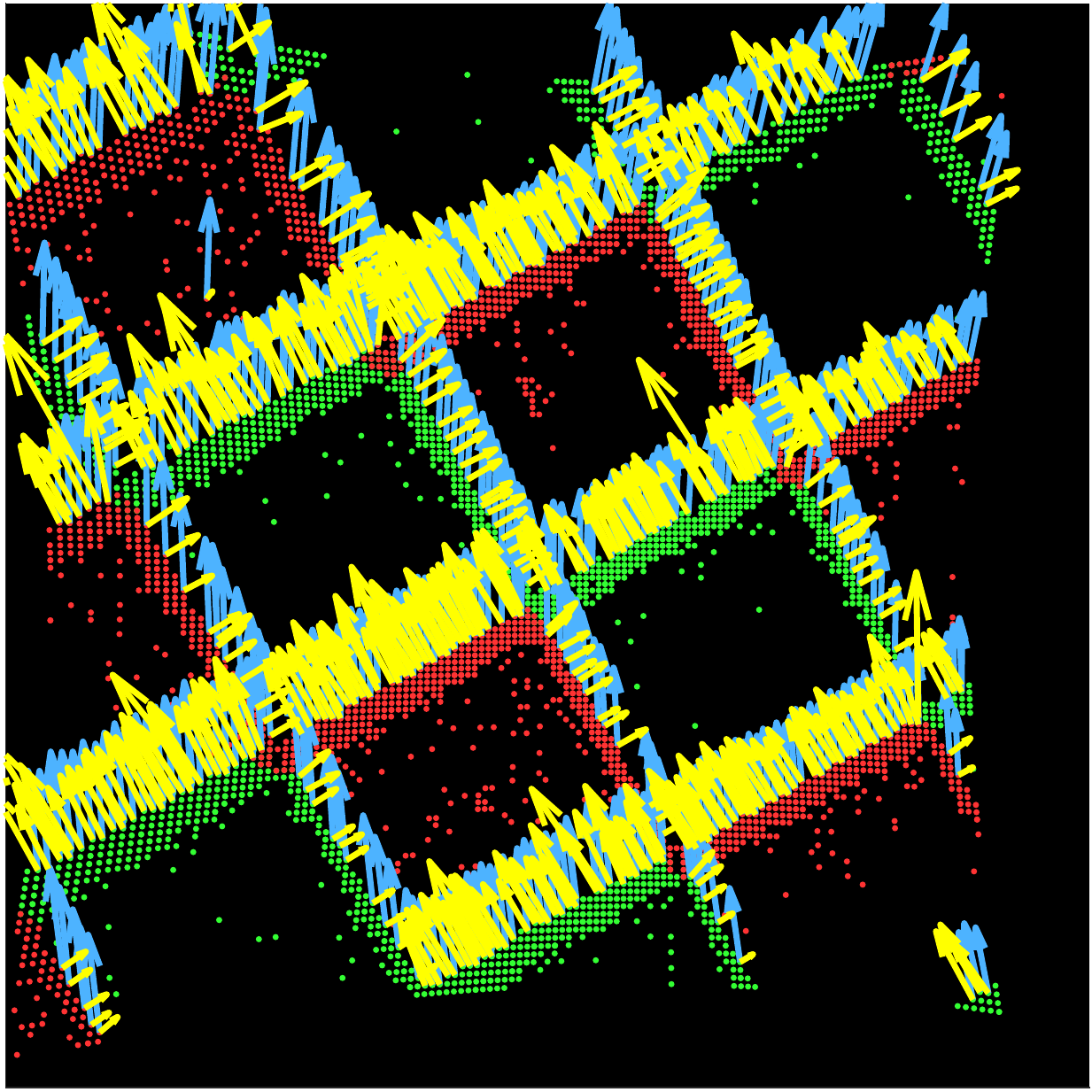}
			\label{fig:flow_checkerboard}
		}
		\subfloat[Rotating checkerboard]{
			\includegraphics[width=0.31\linewidth]{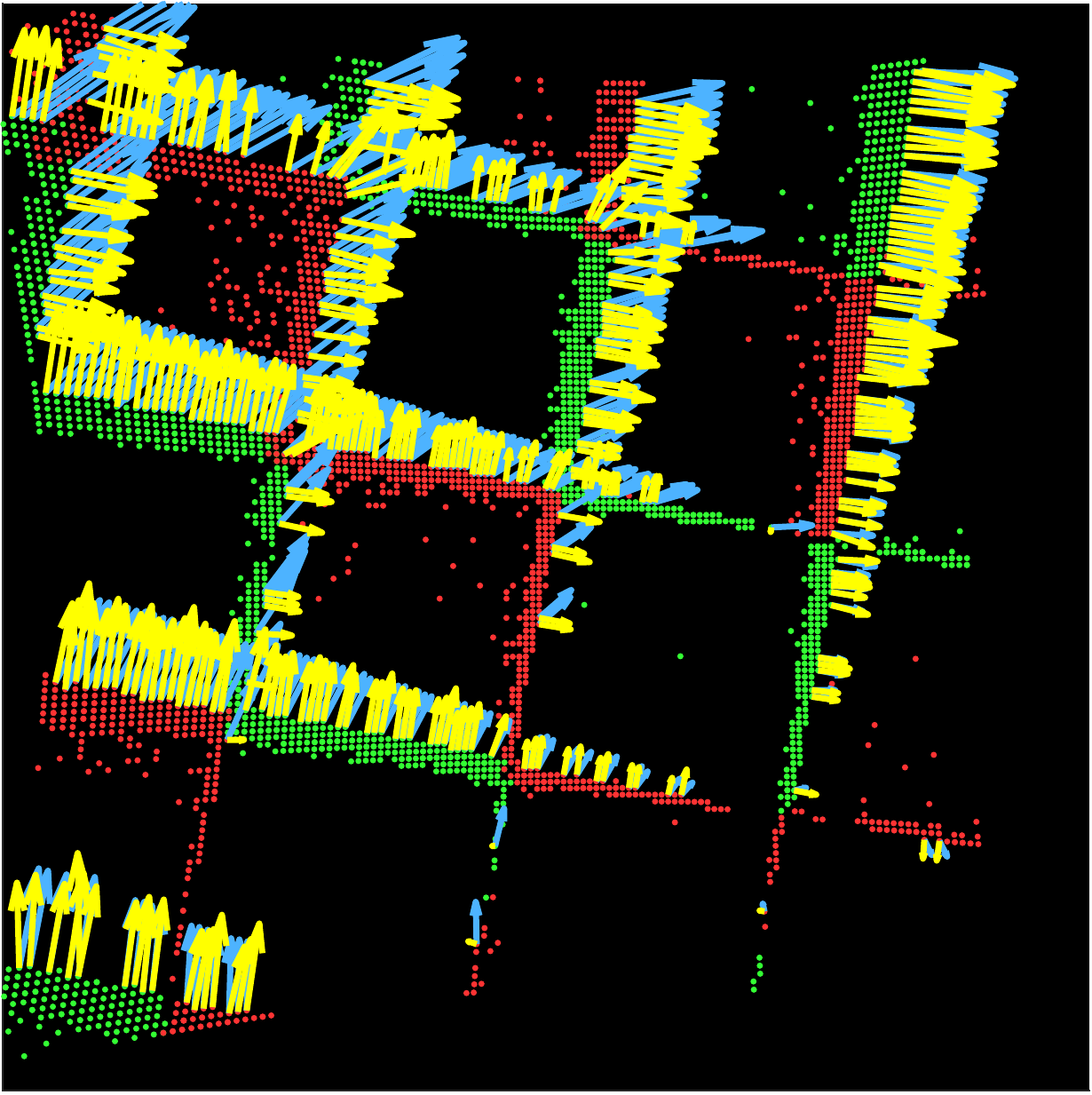}
			\label{fig:flow_rotation}
		}
		\subfloat[Diverging roadmap ($\vartheta_z=1.0$)]{
			\includegraphics[width=0.31\linewidth]{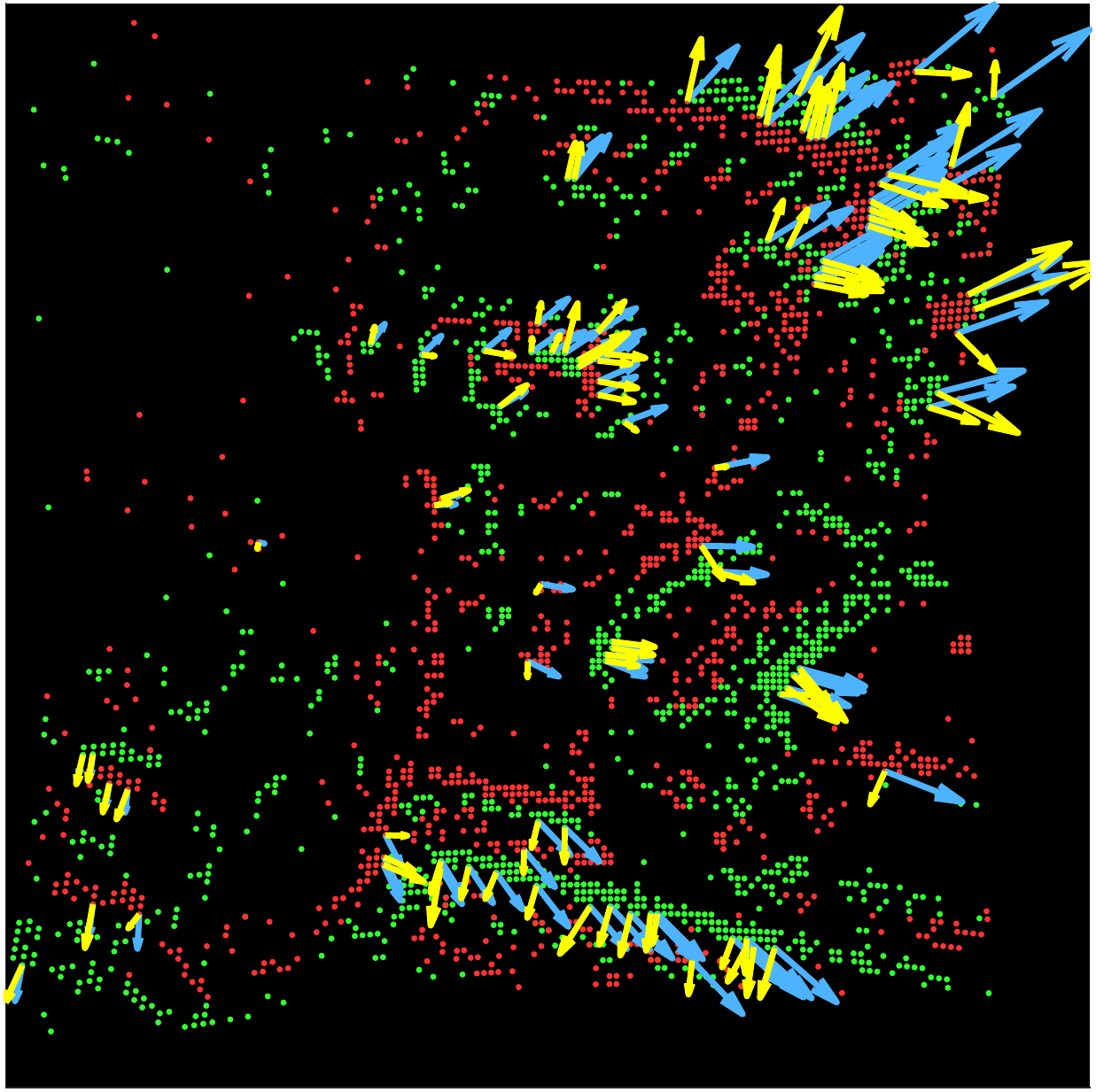}
			\label{fig:flow_roadmap}
		}
		\caption{Optical flow estimated in several sequences, shown as yellow arrows. The accompanying blue arrows show the ground truth optical flow. Events are shown as green dots (positive polarity) or red dots (negative polarity). The time window for displaying optical flow in each sequence is 10 ms. To better visualize the event input, a larger window of 50 ms is applied for the events.}
		\label{fig:flow_results}
	\end{framed}
\end{figure*}

\cref{fig:flow_results} shows optical flow vectors (yellow arrows) estimated using the improved algorithm during three of the selected sequences, along with ground truth flow vectors (blue arrows). Note that, for clarity, the time window for visualizing events is larger than for visualizing optical flow, which is why for some event locations, it appears that no optical flow estimates are found. Accurate normal flow estimates are visible for the checkerboard datasets. In \cref{fig:flow_checkerboard} optical flow is generally constant along the checkerboard edges, matching well to the normal component of the ground truth vectors. The rotating checkerboard sequence (\cref{fig:flow_rotation}) also provides accurate optical flow estimates. Clear variation of the normal flow magnitude along the lines is seen. 

Last, in \cref{fig:flow_roadmap} it is clearly visible that the roadmap texture is more challenging. Event structures are less coherent and the visible features are more noisy. Optical flow vectors are sparsely present, yet the available estimates are sufficient for observing the global motion. At some location with noisy features the motion tends to be underestimated, but the majority of the estimates is very similar to the normal direction of the ground truth motion.

\subsubsection{Quantitative evaluation}
For quantitative evaluation, a comparison was made of the proposed optical flow algorithm and the baseline algorithm by \citet{Benosman2014} (as detailed in \cref{sec:eof_plane_fitting}). In this comparison the fixed time window $\Delta t$ for the original approach is set to 100 ms, and the rejection distance $d_{max}$ is set to 0.001 to obtain a similar event density in both algorithms. For a consistent comparison, the baseline algorithm incorporate the same refractory period $\Delta t_R$, maximum speed limit $V_{max}$, and minimum number of events $n_{min}$ as the proposed algorithm.

For benchmarking optical flow accuracy, several error metrics have been introduced such as the endpoint error and angular error \cite{Baker2011}. These metrics have been incorporated into recent event-based optical flow benchmarks as well \cite{Ruckauer2016}. However, they are defined for optical flow that is fully determinable and is not subject to the aperture problem. The algorithms in this section both estimate normal flow. In \citet{Barranco2014} a version of the endpoint error is applied that indicates the magnitude error of the normal flow with respect to the \emph{projection} of ground truth optical flow along the normal flow vector. This metric is employed here as well. For each optical flow vector, we compute the Projection Endpoint Error (PEE), which is defined as follows:

\begin{equation}
\label{eq:PEE}
\mathrm{PEE} = \left| {\left\| {\bf{V}} \right\| - \frac{{\bf{V}}}{{\left\| {\bf{V}} \right\|}} \cdot {{\bf{V}}_{GT}}} \right|
\end{equation}

where $\mathbf{V}=\left[u,\,v\right]^T$ is the normal flow estimate and $\mathbf{V}_{GT}$ denotes the ground truth optical flow vector.

A comparison of the resulting mean absolute PEE values, and their standard deviations, is presented in \cref{tab:flow_errors}. The optical flow density is also shown (abbreviated here as $\eta$) which indicates the percentage of events for which an optical flow estimate was found. A high value of $\eta$ indicates that more motion information can be obtained with a given event input. 


\begin{table}[!ht]
	\centering
	\setlength{\tabcolsep}{0.5em}
	\renewcommand{\arraystretch}{1.3}
	\caption{Projection Endpoint Error (mean absolute error and standard deviation) and density results of the baseline plane fitting algorithm, and the new algorithm proposed in this work. Values highlighted in bold are the lowest PEE or the highest density result of both algorithms.}
	\footnotesize
	\begin{tabular}{l|cc|cc}
\hline
~ & \multicolumn{2}{c|}{Baseline \cite{Benosman2014}} & \multicolumn{2}{c}{This work} \\ \hline
~ & PEE [pix/s] & $\eta$ [\%] & PEE [pix/s] & $\eta$ [\%] \\ \hline
Checkerboard, $\vartheta_y=1.0$ & 18.6 $\pm$ 22.9 & \textbf{50.6} & \textbf{17.7} $\pm$ 18.7 & 45 \\
Checkerboard, $r=-1.3$ & 26.8 $\pm$ 28.1 & \textbf{50.7} & \textbf{26} $\pm$ 24.7 & 48.2 \\
Checkerboard, $\vartheta_z=0.2$ & \textbf{7.78} $\pm$ 12.1 & 12.6 & 7.81 $\pm$ 8.84 & \textbf{18.8} \\
Checkerboard, $\vartheta_z=0.5$ & 13 $\pm$ 17.9 & 29.5 & \textbf{12.7} $\pm$ 13.9 & \textbf{30.7} \\
Checkerboard, $\vartheta_z=2.0$ & 42.4 $\pm$ 58.4 & \textbf{57.5} & \textbf{36.3} $\pm$ 32.6 & 56.3 \\
Roadmap, $\vartheta_z=0.1$ & \textbf{7.85} $\pm$ 6.91 & 3.54 & 9.73 $\pm$ 8.41 & \textbf{8.93} \\
Roadmap, $\vartheta_z=0.5$ & \textbf{13.5} $\pm$ 13.7 & 8.44 & 13.6 $\pm$ 12.9 & \textbf{14.3} \\
Roadmap, $\vartheta_z=1.0$ & 26.4 $\pm$ 30.3 & 13.3 & \textbf{19.6} $\pm$ 19.6 & \textbf{16.8} \\ \hline

\end{tabular}

	\label{tab:flow_errors}
\end{table} 

Overall, the results are very similar. Both algorithms reach good scores on the checkerboard sets with translation, rotation, and medium divergence ($\vartheta_z=0.5$). However, some differences are observable. The proposed algorithm tends to reach a higher optical flow density in the slow divergence ($\vartheta_z=0.2)$ checkerboard scene and in all roadmap scenes, since the baseline algorithm fails to perceive slowly moving features. Still, this does not degrade the estimate accuracy with respect to the baseline algorithm, or only to a limited extent. Note also that in both fast diverging sequences (Checkerboard, $\vartheta_z=2.0$, and Roadmap, $\vartheta_z=1.0$) a lower mean absolute PEE is achieved with our approach.

\subsubsection{Computational Performance Evaluation}
\label{sec:results_optical_flow_computation}
An assessment of computational complexity is made using two datasets, one for both texture types. Both sets have a duration of 12 s and contain approximately 40k events per second. This enables quantifying the potential of $\rho_{F_{max}}$ to regulate processing time, as well as the effect of texture. The algorithm is implemented in C and interfaced with MATLAB through MEX, running single-threaded on a Windows 10 64bit laptop with an Intel Core i7 Q720 quadcore CPU. Each dataset and setting of $\rho_{F_{max}}$ is processed ten times for consistent results. The CPU usage of MATLAB during the test was around 12\%. 

\begin{figure}[h]
	\centering
	\setlength{\fwidth}{0.5\linewidth}
	\renewcommand{\xlabeldist}{-0.05}
%
%
\definecolor{mycolor1}{rgb}{0.00000,0.44700,0.74100}%
\definecolor{mycolor2}{rgb}{0.85000,0.32500,0.09800}%
\begin{tikzpicture}

\begin{axis}[%
width=\fwidth,
height=0.384\fwidth,
at={(0\fwidth,0\fwidth)},
scale only axis,
xmode=log,
xmin=1000.0000,
xmax=100000.0000,
xlabel={$\rho_{F_{max}}$ [1/s]},
ymin=0.0000,
ymax=2.5000,
ylabel={Computation time [$\upmu$s]},
axis background/.style={fill=white},
legend style={at={(0.97,0.03)},anchor=south east,legend cell align=left,align=left,draw=white!15!black},
title style={font=\labelsize},
xlabel style={font=\labelsize,at={(axis description cs:0.5,\xlabeldist)}},
ylabel style={font=\labelsize,at={(axis description cs:\ylabeldist,0.5)}},
legend style={font=\ticksize},
ticklabel style={font=\ticksize}
]
\addplot [color=mycolor1,solid]
  table[row sep=crcr]{%
1000.0000	0.2098\\
1274.2750	0.2352\\
1623.7767	0.2657\\
2069.1381	0.2997\\
2636.6509	0.3407\\
3359.8183	0.3866\\
4281.3324	0.4365\\
5455.5948	0.4905\\
6951.9280	0.5503\\
8858.6679	0.6121\\
11288.3789	0.6755\\
14384.4989	0.7402\\
18329.8071	0.8031\\
23357.2147	0.8659\\
29763.5144	0.9240\\
37926.9019	0.9807\\
48329.3024	1.0314\\
61584.8211	1.0681\\
78475.9970	1.1148\\
100000.0000	1.1477\\
};
\addlegendentry{Checkerboard};

\addplot [color=mycolor2,solid]
  table[row sep=crcr]{%
1000.0000	0.3599\\
1274.2750	0.4121\\
1623.7767	0.4736\\
2069.1381	0.5387\\
2636.6509	0.6188\\
3359.8183	0.7024\\
4281.3324	0.7924\\
5455.5948	0.8862\\
6951.9280	0.9837\\
8858.6679	1.0839\\
11288.3789	1.1805\\
14384.4989	1.2725\\
18329.8071	1.3605\\
23357.2147	1.4442\\
29763.5144	1.5188\\
37926.9019	1.5865\\
48329.3024	1.6441\\
61584.8211	1.6881\\
78475.9970	1.7335\\
100000.0000	1.7681\\
};
\addlegendentry{Roadmap};

\addplot [color=mycolor1,dashed,forget plot]
  table[row sep=crcr]{%
1000.0000	1.2824\\
100000.0000	1.2824\\
};
\addplot [color=mycolor2,dashed,forget plot]
  table[row sep=crcr]{%
1000.0000	1.8931\\
100000.0000	1.8931\\
};
\end{axis}
\end{tikzpicture}%
	\caption{Processing time per event for checkerboard and roadmap datasets, for different settings of $\rho_{F_{max}}$. The dashed lines indicate the computation times when no limit is applied to $\rho_{F_{max}}$.}
	\label{fig:timing_rate_control}
\end{figure}
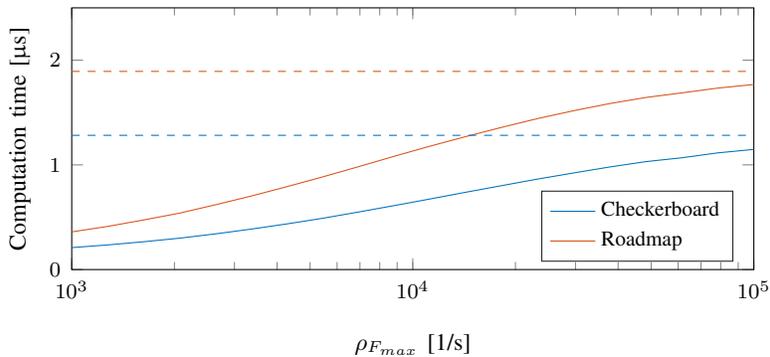

The resulting computation time per event for several settings of $\rho_{F_{max}}$ is shown in \cref{fig:timing_rate_control}, in comparison with the maximal computation time with no control of $\rho_{F_{max}}$. For both textures it is clearly possible to regulate processing time by $\rho_{F_{max}}$. A lower limit appears to be present, which is due to the remaining overhead related to event timestamp copying. Interestingly, there is a clear influence of texture. This difference is due to higher contrast edges in the checkerboard texture, at which several successive events are generated per pixel. Therefore, the refractory period filter rejects these duplicate events before optical flow computation. 

Without the refractory period, computational effort is similar for both textures. In this case, the maximal computation time per event, i.e. without control of $\rho_F$, is 2.11 $\upmu$s. This is equivalent to processing 470k events per second in real-time, which is easily sufficient for processing realistic scenes on the test machine. Event sequences recorded for post-processing contained event peaks below 150k events per second. Nevertheless, if hardware capabilities are more restricted (e.g. in on-board applications), control of $\rho_F$ can be applied to scale the computational complexity of the algorithm down if necessary.

\section{Estimation of Visual Observables from Event-Based Optical Flow}
\label{sec:landing_eof}
This section describes our approach for estimating visual observables from event-based optical flow. While optic flow estimation is performed asynchronously, most existing control systems still operate on a periodic basis. Similarly, the proposed algorithm aims to update the estimates of visual observables at a fixed rate. For each periodic iteration, all newly detected optical flow vectors between the current iteration and the previous one form a planar optical flow field, of which the parameters are estimated. 

The algorithm is based on two components. First, newly detected optical flow vectors are grouped per direction and incorporated into a weighted least-squares estimator for the visual observables, as discussed in \cref{sec:vo_directional_flow_fields}. To enable preservation of flow field information over subsequent periodic iterations, a recursive update technique is introduced in \cref{sec:vo_recursive}. In addition, a confidence value is computed and applied to filter the visual observable estimates, as is described in \cref{sec:vo_confidence}. The estimator is evaluated in combination with our event-based optical flow algorithm in \cref{sec:results_scaled_velocity}.

\subsection{Directional Flow Field Parameter Estimation}
\label{sec:vo_directional_flow_fields}
The presented approach is based on techniques introduced in \citet{DeCroon2013} and used in \citet{Alkowatly2015,Ho2016}, in which fully defined optical flow estimates are available. Since our optical flow algorithm provides normal flow output, a regular optical flow field representation as in \cref{eq:planar_flow_field1} leads to inaccurate parameter estimates. However, in planar flow fields, normal flow may already provide sufficient information for computing the visual observables. Along the direction of the flow vector, normal flow does provide accurate information. 

\begin{figure}[!ht]
	\centering
	\includegraphics[width=0.35\linewidth]{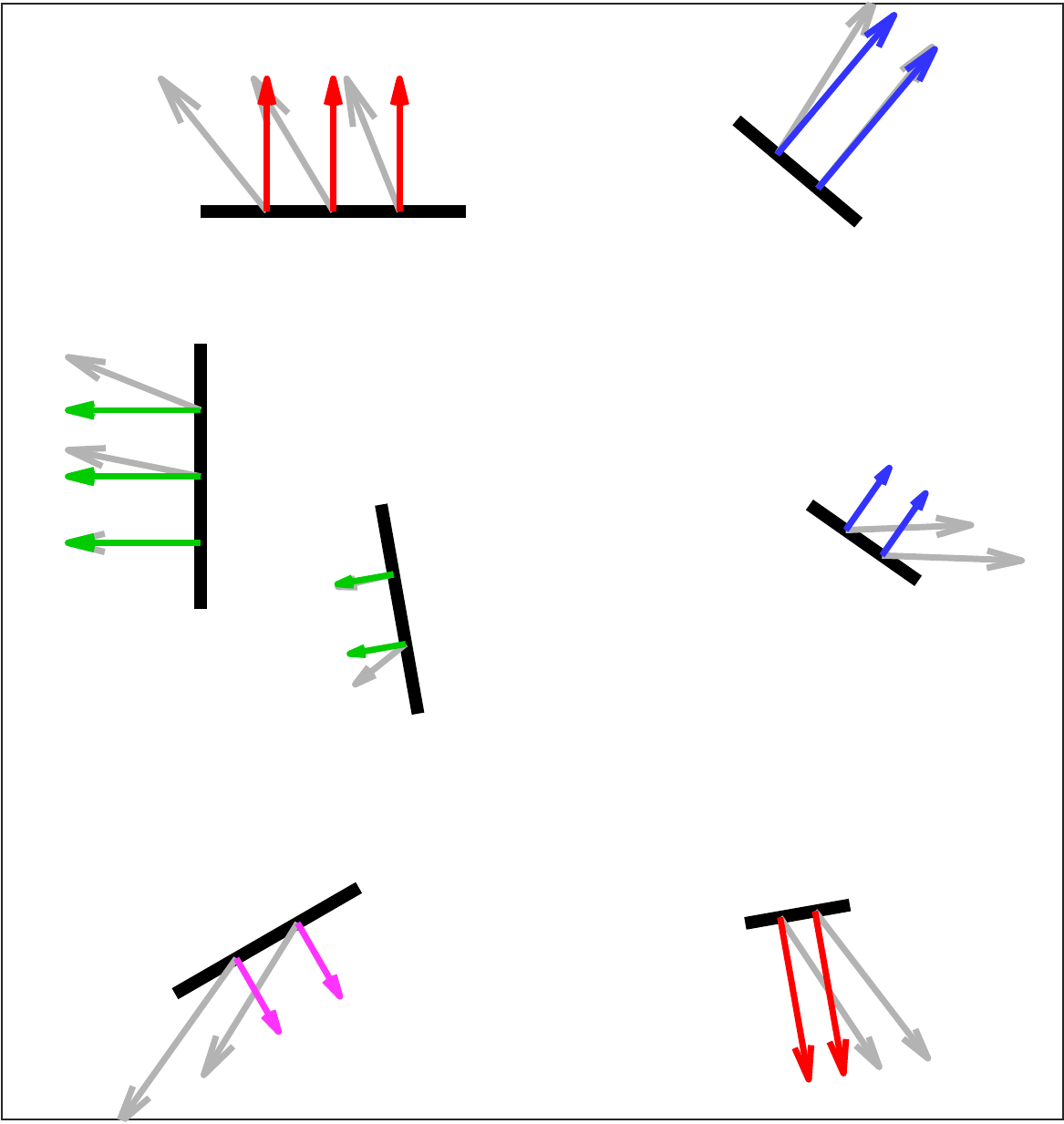}
	\caption{Example of a diverging flow field resulting from several randomly oriented moving edges. The gray vectors indicate the true flow field, while the colored vectors show the normal flow along the edge orientation. Each color indicates a group of normal flow vectors with similar direction.}
	\label{fig:normalFlowField}
\end{figure}

An example diverging flow field with both optical flow and normal flow is sketched in \cref{fig:normalFlowField}. Note that the normal flow in some cases deviates significantly from the optical flow equivalent, which leads to significant errors when computing the flow field parameters. However, when grouped by direction (which is done in \cref{fig:normalFlowField} through the arrow colors), the normal flow vectors indeed show the original pattern of divergence. This idea is central to the proposed directional flow fields approach.

In order to observe flow field divergence along a normal flow direction, at least two separate normal flow vectors are required, whose positions are sufficiently apart. For example, in \cref{fig:normalFlowField} the purple group of normal flow vectors does not, by itself, provide sufficient information for perceiving divergence. Also, if the flow vectors are located in close proximity, errors in normal flow magnitude have a larger influence. In \cref{fig:normalFlowField} the green group is more sensitive to these errors than the red group, since the edges are located closely together. Grouping per direction enables assessment of the reliability of the flow field in each direction, taking the previous issues into account.

A set of $m$ directions $\lbrace\alpha_1,\alpha_2,\ldots,\alpha_N\rbrace$ is defined, where $\alpha_1=0$ and $\alpha_i-\alpha_{i-1} = \pi/m$. In this work, $m=6$ directions are used. For each newly available flow vector, we first determine the closest match of $\alpha_i$ to the flow direction $\alpha_f$. Each direction $\alpha_i$ accommodates both flow in similar and opposite direction, i.e. when $-\pi<\alpha_f<0$, a match is computed for $\alpha_f + \pi$. 

Along the selected direction $\alpha_i$, the projected normal flow position $S$ and magnitude $V$ are computed, hence obtaining a one-dimensional representation of the flow along $\alpha_i$:

\begin{equation}
\label{eq:transform_flow_field}
\left[ {\begin{array}{*{20}{c}}
	S\\
	V
	\end{array}} \right] = \left[ {\begin{array}{*{20}{c}}
	{\hat x}&{\hat y}\\
	{\hat u}&{\hat v}
	\end{array}} \right]\left[ {\begin{array}{*{20}{c}}
	{\cos \alpha_i }\\
	{\sin \alpha_i }
	\end{array}} \right]
\end{equation}

Subsequently, it is corrected for rotational motion by subtracting the normal component of the rotational flow:

\begin{equation}
\begin{aligned}
{V_T} = V &- \cos {\alpha _i}\left( {p - \hat yr - q\hat x\hat y + p{{\hat x}^2}} \right) \\~&+ \sin {\alpha _i}\left( {q - \hat xr - p\hat x\hat y + q{{\hat y}^2}} \right)
\end{aligned}
\end{equation}

For each direction, a one-dimensional flow field is maintained. From \cref{eq:planar_flow_field1} and \cref{eq:transform_flow_field}, the flow field in a single direction is expressed as:

\begin{equation}
\label{eq:flow_field_line}
V_T =  - {\vartheta _x}\cos \alpha_i  - {\vartheta _y}\sin \alpha_i  + {\vartheta _z}S
\end{equation}

To solve \cref{eq:flow_field_line} for the visual observables, a weighted least-squares solution is computed using the flow vectors from all directions. Let $\mathrm{c}_\alpha=\cos\alpha$ and $\mathrm{s}_\alpha=\sin\alpha$. The overdetermined system to be solved is composed as follows:
\begin{equation}
\label{eq:dir_flow_field_system}
\arraycolsep=1.4pt
\left[ {\begin{array}{*{20}{c}}
	{ - {\rm{c}}_{\alpha _1}}&{ - {\rm{s}}_{\alpha _1}}&{{S_{1,1}}}\\
	\vdots & \vdots & \vdots \\
	{ - {\rm{c}}_{\alpha _1}}&{ - {\rm{s}}_{\alpha _1}}&{{S_{1,{n_1}}}}\\
	{ - {\rm{c}}_{\alpha _2}}&{ - {\rm{s}}_{\alpha _2}}&{{S_{2,1}}}\\
	\vdots & \vdots & \vdots \\
	{ - {\rm{c}}_{\alpha _2}}&{ - {\rm{s}}_{\alpha _2}}&{{S_{1,{n_2}}}}\\
	\vdots & \vdots & \vdots \\
	{ - {\rm{c}}_{\alpha _m}}&{ - {\rm{s}}_{\alpha _m}}&{{S_{m,{n_m}}}}
	\end{array}} \right]\left[ {\begin{array}{*{20}{c}}
	{{\vartheta _x}}\\
	{{\vartheta _y}}\\
	{{\vartheta _z}}
	\end{array}} \right] \approx \left[ {\begin{array}{*{20}{c}}
	{{V_{1,1}}}\\
	\vdots \\
	{{V_{1,{n_1}}}}\\
	{{V_{2,1}}}\\
	\vdots \\
	{{V_{2,{n_2}}}}\\
	\vdots \\
	{{V_{m,{n_m}}}}
	\end{array}} \right]
\end{equation}
which has the form $\mathbf{A\Theta}\approx\mathbf{y}$. The weighted least-squares solution is then obtained from the normal equations:

\begin{equation}
\label{eq:weighted_least_squares}
\mathbf{A}^T\mathbf{W}\mathbf{A}\mathbf{\Theta}=\mathbf{A}^T \mathbf{Wy}
\end{equation} 
in which $\mathbf{W}$ a diagonal matrix composed of the weights per direction:

\begin{equation}
\mathbf{W} = \mathrm{diag}\Big(W_1,\cdots,W_1,W_2,\cdots,W_2,\cdots, W_m\Big)
\end{equation}

The weight $W_i$ is used to represent the reliability of normal flow along a direction $i$ based on the spread of $S_i$ along that direction. Its value is determined by the variance $\mathrm{Var}\lbrace S_i\rbrace$. We let $W_i$ scale linearly with $\mathrm{Var}\lbrace S_i\rbrace$, up to a maximum of $\mathrm{Var}\lbrace S\rbrace_{min}$:

\begin{equation}
\label{eq:weight_variance}
{W_{i} = \left\{ {\begin{array}{*{20}{c}}
	0&{\mathrm{Var}\lbrace S_i\rbrace= 0}\\
	{\frac{\mathrm{Var}\lbrace S_i\rbrace}{\mathrm{Var}\lbrace S\rbrace_{min}}}&0<{{\mathrm{Var}\lbrace S_i\rbrace} \le {\mathrm{Var}\lbrace S\rbrace_{min}}}\\
	1&{{\mathrm{Var}\lbrace S_i\rbrace} > {\mathrm{Var}\lbrace S\rbrace_{min}}}
	\end{array}} \right.}
\end{equation}

The minimum variance $\mathrm{Var}\lbrace S\rbrace_{min}$ is set to 600 pixels$^2$. 

Note also that, through the formulation of \cref{eq:dir_flow_field_system}, directions with more normal flow estimates have a larger influence on $\mathbf{\Theta}$. Hence, directions for which more information is available, contribute more to the solution.

\subsection{Recursive Updating of the Flow Field}
\label{sec:vo_recursive}
The solution to \cref{eq:weighted_least_squares} for $\mathbf{\Theta}$ provides the estimate for the visual observables. However, depending on the sampling rate of the estimator, it is possible that, during a single periodic iteration, too few normal flow estimates are available for an accurate fit. This leads to noise peaks in the measurement of $\mathbf{\Theta}$, especially during low speed motion. To limit this effect, the matrices $\mathbf{A}$ and $\mathbf{y}$ are not completely renewed at each iteration. Instead, rows from previous iterations are retained and assigned an exponentially decreasing weight, similar to an exponential moving average filter.

For an efficient implementation of the former, $\mathbf{A}$ and $\mathbf{y}$ are not explicitly composed as shown in \cref{eq:dir_flow_field_system}. Instead, our approach operates on the normal equations in \cref{eq:weighted_least_squares}. For each direction independently, we recursively update parts of the matrices $\mathbf{B}={{\bf{A}}^T}{\bf{WA}}$ and $\mathbf{C}={{\bf{A}}^T}{\bf{Wy}}$. These matrices are composed by the following elements:

\begin{equation}
\bf{B} = \left[ {\begin{array}{*{20}{c}}
	{{b_{11}}}&{{b_{21}}}&{{b_{31}}}\\
	{{b_{21}}}&{{b_{22}}}&{{b_{32}}}\\
	{{b_{31}}}&{{b_{32}}}&{{b_{33}}}
	\end{array}} \right],\;\mathbf{C} = \left[ {\begin{array}{*{20}{c}}
	{{c_1}}\\
	{{c_2}}\\
	{{c_3}}
	\end{array}} \right]
\end{equation}

From \cref{eq:dir_flow_field_system}, it can be shown that the elements of $\mathbf{B}$ are expressed as:

\begin{equation}
\def\arraystretch{2.2}
\begin{array}{c@{\hspace{0.5em}}c@{\hspace{0.5em}}c@{\hspace{0.5em}}c@{\hspace{0.5em}}c@{\hspace{0.5em}}c@{\hspace{0.5em}}c@{\hspace{0.5em}}}
{b_{11}} &=& \sum\limits_{i=1}^m {{W_i}{n_i}{{\left( {{\rm{c}}_{\alpha _i}} \right)}^2}}, &\;& {b_{21}} &=& \sum\limits_{i=1}^m {{W_i}{n_i}{\rm{c}}_{\alpha _i}{\rm{s}}_{\alpha _i}} \\
{b_{22}} &=& \sum\limits_{i=1}^m {{W_i}{n_i}{{\left( {{\rm{s}}_{\alpha _i}} \right)}^2}}, &\;& {b_{31}} &=& \sum\limits_{i=1}^m {{W_i}{\rm{c}}_{\alpha _i}\sum\limits_{j=1}^{n_i} {S_{i,j}}}  \\
{b_{33}} &=& \sum\limits_{i=1}^m {{W_i}\sum\limits_{j=1}^{n_i} {S_{i,j}^2} }, &\;&{b_{32}} &=& \sum\limits_{i=1}^m {{W_i}{\rm{s}}_{\alpha _i}\sum\limits_{j=1}^{n_i} {{S_{i,j}}}} 
\end{array}
\end{equation}

and those of $\mathbf{C}$ are expressed as:

\begin{equation}
\def\arraystretch{2.2}
\begin{array}{c@{\hspace{0.5em}}c@{\hspace{0.5em}}c@{\hspace{0.5em}}}
{c_1} &=& \sum\limits_{i=1}^m {{W_i}{\rm{c}}_{\alpha _i}\sum\limits_{j=1}^{n_i} {{V_{i,j}}} }\\
{c_2} &=& \sum\limits_{i=1}^m {{W_i}{\rm{s}}_{\alpha _i}\sum\limits_{j=1}^{n_i} {{V_{i,j}}} }\\
{c_3} &=& \sum\limits_{i=1}^m {{W_i}\sum\limits_{j=1}^{n_i} {{S_{i,j}}{V_{i,j}}} } 
\end{array}
\end{equation}

We introduce a shorthand notation $\Sigma_S^i = \sum_{j=1}^{n_i}S_{i,j}$ to represent the sums, cross-product sums, and sums of squares of $S$ and $V$ for direction $i$. The unweighted contribution of the associated flow vectors is then contained in $n_i$ and the sums $\Sigma_S^i$, $\Sigma_{S^2}^i$, $\Sigma_{V}^i$, and $\Sigma_{SV}^i$. These values are further referred to as the \emph{flow field statistics}. Hence, a newly detected flow vector is included in the flow field estimate by incrementing these quantities according to the values $S$ and $V$ of the new vector. 

What makes this decomposition interesting, is that the flow field statistics form a compact summary of the flow field, independent of the actual number of flow vectors. Thus, flow field information from a previous iteration can be efficiently included in subsequent ones, without increasing the size of the system in \cref{eq:dir_flow_field_system}. Now, at the start of each iteration, it is possible to include information from the flow field of the previous iteration, simply by preserving a fraction $F$ of the previous flow field statistics. Hence, the estimator accuracy is less dependent on the sampling rate of the algorithm.

The preservation process is illustrated using the statistic $\Sigma_{V}^i$. At the start of iteration $k$, $\Sigma_{V}^i$ is initialized as $\Sigma_{V}^i(k) = F \Sigma_{V}^i(k-1)$. During iteration $k$, $\Sigma_{V}^i$ is then updated using newly available normal flow vectors that are allocated to direction $i$. Hence, the complete update for $\Sigma_{V}^i$ is performed as follows:
\begin{equation}
\Sigma_{V}^i(k) = F \Sigma_{V}^i(k-1) + \sum_{j=1}^{n_i}{V_{i,j}}
\end{equation}

The value of $F$ is computed as:

\begin{equation}
F = 1-\frac{t(k)-t(k-1)}{k_f}
\end{equation}

where the time constant $k_f$ is assigned a value of 0.02 s. This step is similar for all statistics. When all newly available vectors are categorized and processed, the flow field is recomputed using \cref{eq:weighted_least_squares}.

%
%

\subsection{Confidence Estimation and Filtering}
\label{sec:vo_confidence}
In visual sensing, the reliability of motion estimates varies greatly depending on the environment. Factors such as visible texture and scene illumination have an effect on the estimate. With event-based sensing, motion in the scene is another key factor. 

Therefore, a confidence value is computed based on several characteristics of the flow field, in order to quantify the reliability of the estimate. This confidence value is defined as a product of three individual confidence metrics based on the following statistical quantities: 

\begin{itemize}
	\item The flow estimation rate $\rho_F$.
	\item The maximal variance $\mathrm{Var}\lbrace S\rbrace$ of all flow directions.
	\item The coefficient of determination $R^2$ of the solution to \cref{eq:weighted_least_squares}, applied here as a nondimensional measure of the fit quality.
\end{itemize}

$R^2$ is generally computed through the following \cite{Weisberg2005}:

\begin{equation}
\label{eq:R2}
{R^2} = 1 - \frac{\mathit{RSS}}{\mathit{TSS}}
\end{equation}

In this work, the Residual Sum of Squares (RSS) and Total Sum of Squares (TSS) are computed in weighted form as follows:

\begin{equation}
\begin{aligned}
\mathit{RSS} &=& {{\bf{y}}^T}{\bf{Wy}} - {{\bf{\Theta }}^T}{{\bf{A}}^T}{\bf{Wy}}\\
\mathit{TSS} &=& {{\bf{y}}^T}{\bf{Wy}} - \frac{\left({\sum\limits_{i = 1}^m {W\Sigma _V^i} }\right)^2}{{\sum\limits_{i = 1}^m {W{n_i}} }}
\end{aligned}
\end{equation}

For each indicator, a confidence value $k$ is computed ranging from 0 to 1 (higher is better), similar to the variance weight in \cref{eq:weight_variance}. The individual confidence values are thus dependent on settings for $R^2_{min}$, $\mathrm{Var}\lbrace S\rbrace_{min}$, and $\rho_{F_{min}}$ (not to be confused with $\rho_{F_{max}}$). The values of $R^2_{min}$ and $\rho_{F_{min}}$ are set to 1.0 and 500 respectively. Note that, since \cref{eq:weight_variance} already provides individual confidence values per direction in the form of $W$, we simply let $k_{\mathrm{Var}\lbrace S\rbrace}=\mathrm{max}\left({W_i:i=1,\ldots,m}\right)$.

The total confidence value $K$ is then the product of $k_{\rho_F}$, $k_{\mathrm{Var}\lbrace S\rbrace}$, and $k_{R^2}$. Hence, each individual confidence factor needs to be close to 1 in order to obtain a high $K$. For example, when $\rho_F$ and $R^2$ are very large, but the flow is very localized (the maximal value for $\mathrm{Var}\lbrace S\rbrace$ is small), the estimate is still not reliable. In this case, it is likely that a single visual feature causes the normal flow, which is insufficient for computing the visual observables.

The confidence $K$ is useful to monitor the estimate quality of the visual observables during flight. In addition, it is the main component of a \emph{confidence filter} for $\mathbf{\Theta}$. This filter is based on a conventional infinite impulse response low-pass filter, in which $K$ is multiplied with the filter's update constant. The final estimate for the visual observables $\hat{\mathbf{\Theta}}$ is determined through the following update equation at iteration $k$:

\begin{equation}
\mathbf{\hat\Theta} (k)=\mathbf{\hat\Theta}(k-1) + \left(\mathbf{\Theta}(k) - \mathbf{\hat\Theta}(k-1)\right)K \frac{t(k)- t(k-1)}{k_t}
\end{equation}

where $k_t$ is the time constant of the low-pass filter, which is set to 0.02 s. Lastly, a saturation limit is applied that caps the magnitude of the update of each individual value in $\mathbf{\Theta}$ to $\Delta \vartheta_{max}$ in order to reject significant outliers. The value for $\Delta \vartheta_{max}$ is set to 0.3.

\subsection{Results}
\label{sec:results_scaled_velocity}
For evaluating the accuracy of the presented visual observable estimator, we use the measurements generated for evaluating optical flow performance in \cref{sec:results_optical_flow}, which are generated through handheld motion. Optitrack position measurements provide the ground truth estimates for $\vartheta_x$, $\vartheta_y$, and $\vartheta_z$. For each set, normal flow estimates are computed using the C-based implementation discussed in \cref{sec:results_optical_flow}. The flow detection rate cap $\rho_{F_{max}}$ is set to 2500 flow vectors per second and the periodic estimator samples the visual observables at 100 Hz, similar to the on-board implementation in \cref{sec:results}.

In our experiments the main variable of interest is $\vartheta_z$, as it forms the basis for the constant divergence controller. Therefore, this variable is investigated over a wide range of velocities. However, the estimates of $\vartheta_x$ and $\vartheta_y$ are also interesting to assess, since a more elaborate optical flow based controller may also include the horizontal components for hover stabilization. The latter process does require the MAV to perform rolling and pitching motion, inducing rotational normal flow. Therefore, the effectiveness of derotation is evaluated as well.

\subsubsection{Vertical Motion}
For assessment of $\vartheta_z$ estimates, vertical oscillating motion was performed above both texture types. The vertical speed of these oscillations was gradually increased, hence covering a wide range of divergence values. This enables a first-order characterization of the estimator behavior.

\cref{fig:divergence_estimates} shows the resulting estimates compared to ground truth measurements, accompanied by height measurements $h=-Z_{\cal W}$. Detail sections are shown for low and high divergence motion, for which also the confidence value is shown. 

\begin{figure*}[!ht]
	\centering
	\setlength{\fwidth}{0.4\linewidth}
	\begin{framed}
		\subfloat[Checkerboard texture]	{
			\input{images/divergence_checkerboard_main}
			\label{fig:divergence_checkerboard}
		}
		\subfloat[Roadmap texture]	{
			\input{images/divergence_roadmap_main}
			\label{fig:divergence_roadmap}
		}	
		\caption{From top to bottom: Height measurements (top row) and estimates of $\vartheta_z$ (second row, red line) in comparison to ground truth measurements (blue line). The two bottom rows show detail sections of $\vartheta_z$ estimates (third row) at low speed and high speed, as well as the accompanying estimate confidence value $K$ (bottom row). Measurements are shown for \protect\subref{fig:divergence_checkerboard} checkerboard and \protect\subref{fig:divergence_roadmap} roadmap textures separately.}
		\label{fig:divergence_estimates}
	\end{framed}
\end{figure*}

The detail plots show that the estimator is relatively sensitive to local outliers in normal flow at low speeds. In addition, the confidence $K$ is generally low due to lower detection rates of optical flow and low value of $R^2$. At higher speeds, the errors are relatively smaller. Note that $K$ is also generally higher there. Somewhat lower confidence values are seen for the roadmap texture.

Around sign changes, brief moments are present where the confidence value $K$ is low. The result of this is that, due to the confidence filter, the update of $\hat{\vartheta}_z$ at these points is limited, which leads to a local delay with respect to the ground truth. However, when higher confidence estimates are available, the estimate quickly converges back to the ground truth value.

Based on the estimator results in \cref{fig:divergence_estimates} we can assess how the error varies with the ground truth divergence. \cref{fig:divergence_error_dist} shows the variation of the absolute error $\varepsilon_{\vartheta_z}=\lvert\hat{\vartheta}_z-\vartheta_z \rvert$ with the magnitude of $\vartheta_z$\footnote{The dataset used to generate these plots and statistics is currently being prepared to be made public. They are planned to be made public at: \url{https://beta.dataverse.nl/dataverse/mavlab}}. A quadratic model $\varepsilon = p_0 + p_1 \vartheta_z + p_2 \vartheta_z^2$ is fitted to the points, which is represented by the blue line. The values of $p_0$, $p_1$, and $p_2$ are shown in \cref{tab:div_error_parameters}. The errors of both the checkerboard set and the roadmap set are combined, since the estimator shows roughly the same error distribution for both cases with the quadratic fit almost flat for the divergences tested. Interestingly, the largest absolute errors appear to be present at low divergence. This results from the local delay occurring around zero-crossings in \cref{fig:divergence_estimates} where the confidence value of the filter is low. Note, however, that the error increase with the magnitude of $\lvert\vartheta_z\rvert$ is limited, which enables application of the presented pipeline to a wide range of velocities.

\begin{figure}[!ht]
	\centering
	\setlength{\fwidth}{0.6\linewidth}
	\renewcommand{\xlabeldist}{0.0}
	\renewcommand{\ylabeldist}{0.0}
	\input{images/divergence_abs_error}
	\caption{Absolute error distribution for the estimates of $\vartheta_z$ for a set of landing tests performed at different constant vertical speeds above the roadmap texture. The blue line shows a quadratic fit of the error and the dashed black line shows the 25, 50 and 75\% percentiles of the data. For comparison, the model obtained for the frame-based size divergence estimator \cite{Ho2016a} is shown as well.} 
	\label{fig:divergence_error_dist}
\end{figure}

In \citet{Ho2016a} an extensive characterization of two frame-based visual estimators for $\vartheta_z$ is performed, which includes an assessment of their absolute error distribution up to $\vartheta_z\approx 1.3$ (Fig. 10 in the paper). For a first-order comparison, \cref{fig:divergence_estimates} also shows the quadratic error fit obtained for the frame-based 'size divergence' estimator, which performed best in \citet{Ho2016a}. Compared to the presented event-based estimator, the size divergence estimator achieves slightly lower errors in the region of $\vartheta_z < 0.5$. However, for faster motion, the error is lower for our event-based estimator. Note that our quadratic model is based on relatively little measurements, and does not yet provide a full characterization.

\begin{table}[!ht]
	\centering
	\caption{Parameters of the quadratic fit models in \cref{fig:divergence_error_dist}.}
	\begin{tabular}{l|cc}
\hline
~ & This work & Size divergence \cite{Ho2016a} \\ \hline
$p_0$ & 0.0359 & 0.0455 \\
$p_1$ & -0.0012 & -0.0043 \\
$p_2$ & 0.0468 & 0.1841 \\ \hline

\end{tabular}

	\label{tab:div_error_parameters}
\end{table}

\subsubsection{Horizontal Motion}
Estimation performance for the components $\vartheta_x$ and $\vartheta_y$ is assessed through a dataset consisting of primarily horizontal motion. In the following set, the DVS is moved in a circular pattern above a checkerboard surface at approximately 0.8 m height.

\begin{figure}[!ht]
	\centering
	\setlength{\fwidth}{0.4\linewidth}
	\input{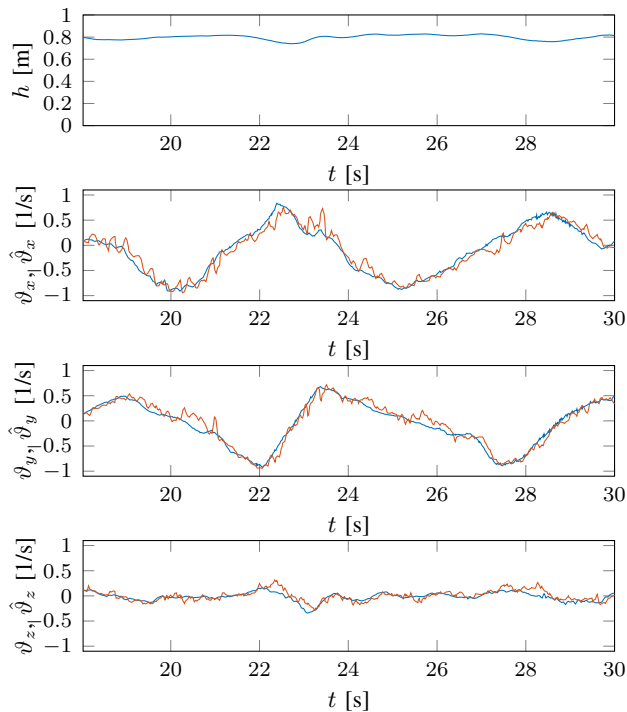}
	\caption{Height and visual observable measurements (blue) and estimates (red) during horizontal motion above checkerboard texture.}
	\label{fig:horizontal_motion}
\end{figure}

The resulting visual observable estimates are shown in \cref{fig:horizontal_motion}. For completeness, the values of $\vartheta_z$ are displayed as well. Overall, the horizontal movement is captured well in the estimates, although some disturbances are still clearly present, for example around $t=23$ s. The deviations are comparable to those seen in the vertical motion dataset. A summary of the error values is presented in \cref{tab:horizontal_errors}.

\begin{table}[!ht]
	\centering
	\caption{Mean and standard deviation of absolute errors for the estimates during horizontal motion, shown in \cref{fig:horizontal_motion}.}
	\begin{tabular}{l|cc}
\hline
~ & Mean abs. error [1/s] & Standard deviation [1/s] \\ \hline
$\vartheta_x$ & 0.09997 & 0.081676 \\
$\vartheta_y$ & 0.077126 & 0.066433 \\
$\vartheta_z$ & 0.051617 & 0.047676 \\ \hline

\end{tabular}

	\label{tab:horizontal_errors}
\end{table}

\subsubsection{Effect of Derotation}
\begin{figure}[!ht]
	\centering
  	\setlength{\fwidth}{0.4\linewidth}
	\input{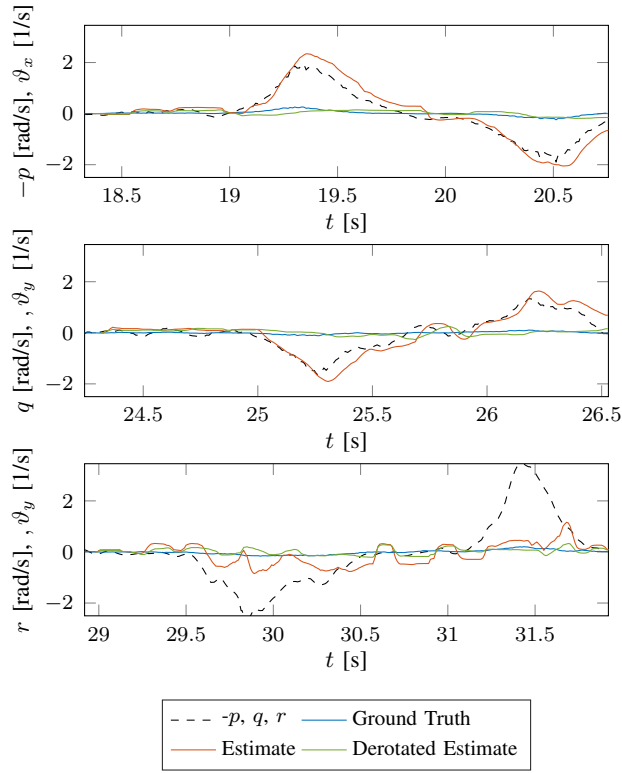}
	\caption{Baseline and derotated estimates of $\vartheta_x$, $\vartheta_y$ compared to ground truth measurements and body rates $p$, $q$, and $r$. Note that the sign of $p$ is inverted to match $\vartheta_x$.}
	\label{fig:derotation}
\end{figure}

In order to assess how well normal flow can be derotated with the current setup, measurements were conducted in which the DVS performed pure rotation along all axes. We compare body rate measurements, ground truth values for $\vartheta_x$ and $\vartheta_y$, and estimates with and without derotation. \cref{fig:derotation} shows these quantities obtained in three separate sequences, in which each body rate is varied independently. 

The most relevant influences are those of $p$ on $\vartheta_x$ and $q$ on $\vartheta_y$, which are shown in the top and middle graphs respectively. The influence of $r$ is less profound. For conciseness, only the effect of $r$ on $\vartheta_y$ is shown, which provided the clearest result. Some residual motion in $p$ and $q$ is present in the latter case, though it does not fully account for the deviation seen in \cref{fig:derotation}. Note that the derotation process generally performs well; the largest part of the rotational flow is successfully corrected in the derotated estimate. 

\section{Constant Divergence Landing Experiments}
\label{sec:results}
This section presents experimental results of constant divergence landings with the presented algorithms in the control loop. In \cref{sec:results_controller} the divergence control law is defined, after which the experimental setup is detailed in \cref{sec:experimental_setup}. Results from the experiments are presented and discussed in \cref{sec:results_div_landings}.

\subsection{Divergence Controller}
\label{sec:results_controller}
The control law regulates $\vartheta_z$ through the vertical thrust $T$. The controller applies a thrust difference $\Delta T$ with respect to a nominal hover thrust $T_0$, such that $T = T_0 + \Delta T$. A simple proportional control law is applied to $\Delta T$ based on $\vartheta_z$, similar to \citet{DeCroon2016}:

\begin{equation}
\Delta T = k_P \left(\vartheta_{z_{r}}-\vartheta_z\right)
\end{equation}

The nominal hover thrust $T_0$ counteracts the weight of the test vehicle. Its value is adapted in-flight in the height control loop of the test vehicle's autopilot software. Before the start of each landing, the vehicle first performs automatic hover to obtain a stable estimate for $T_0$. During the subsequent landing maneuver its value is kept constant.

\subsection{Experimental Setup}
\label{sec:experimental_setup}
The flying platform used in this work is a customized quadrotor referred to as the MavTec. Its main component is a Lisa/M board, which features a 72MHz 32bit ARM microprocessor as well as a pressure sensor and 3-axis rate gyros, accelerometers, and magnetometers. The Lisa/M runs the open-source autopilot software Paparazzi\footnote{Paparazzi UAV, \url{http://wiki.paparazziuav.org/}}, which handles the control of the drone\footnote{Code used in project is publicly available at: \url{https://github.com/tudelft/paparazzi/tree/event_based_flow}}. The DVS is mounted at the bottom of the MavTec facing downwards, aligned according to the reference frame definitions of $\cal C$ and $\cal B$ in \cref{sec:model}. Experiments are performed indoors, using an Optitrack motion tracking system to measure ground truth position and attitude.

In addition, an Odroid XU4 board is mounted on the quadrotor, which processes the event output of the DVS. It features a Samsung Exynos 5422 octacore CPU (four cores at 2.1 GHz and four at 1.5 GHz). The Odroid receives the events from the DVS through a USB 2.0 connection and processes these through the C-based open-source software cAER\footnote{Code used in project is publicly available at: \url{https://github.com/tudelft/caer/tree/odroid-dvs}} \cite{Longinotti2014}. 


\begin{figure*}[!ht]
	\centering
	\begin{framed}
		\begin{minipage}[c]{0.3\textwidth}
			\centering
		\subfloat[Top view]{
			\includegraphics[width=\textwidth]{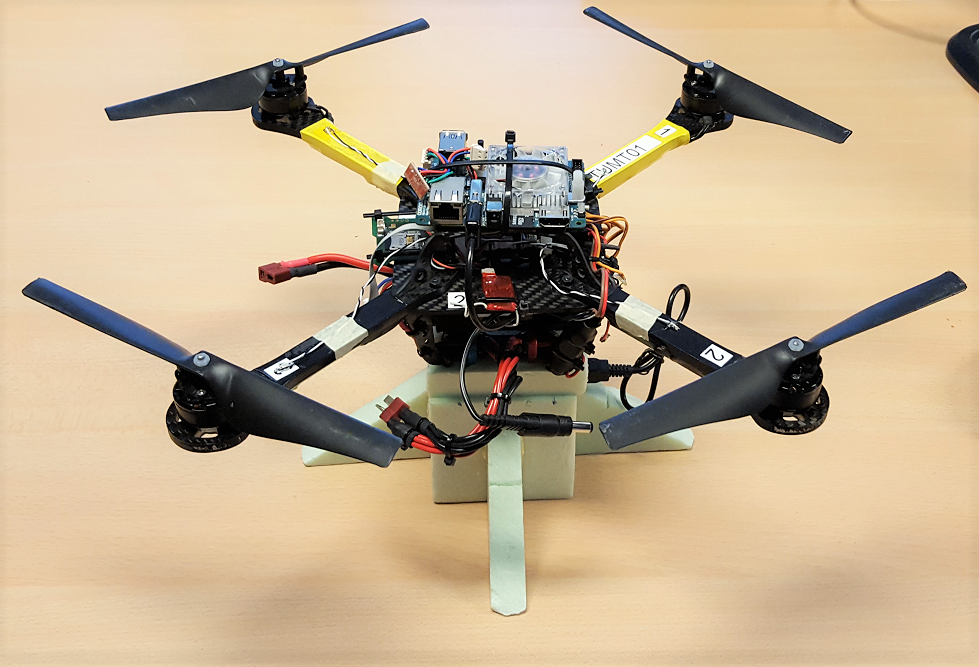}
			\label{fig:mavtec_top}
		}\par\vfill
		\subfloat[Bottom view showing the DVS]{
			\includegraphics[width=\textwidth]{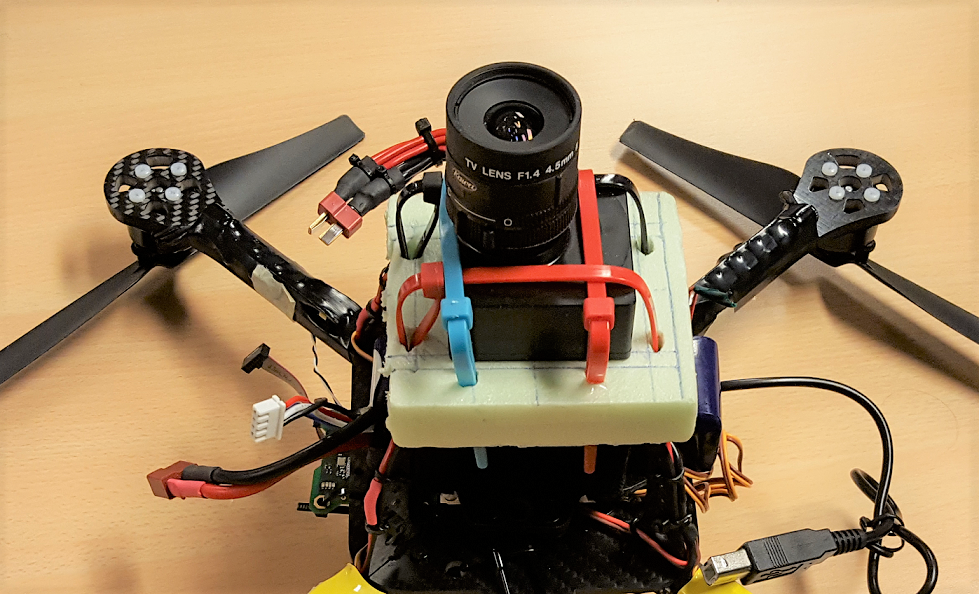}
			\label{fig:mavtec_bottom}
		}
		\end{minipage}\hfill
		\begin{minipage}[c]{0.67\textwidth}
			\subfloat[Overview of the implementation]{
				\includegraphics[width=\textwidth]{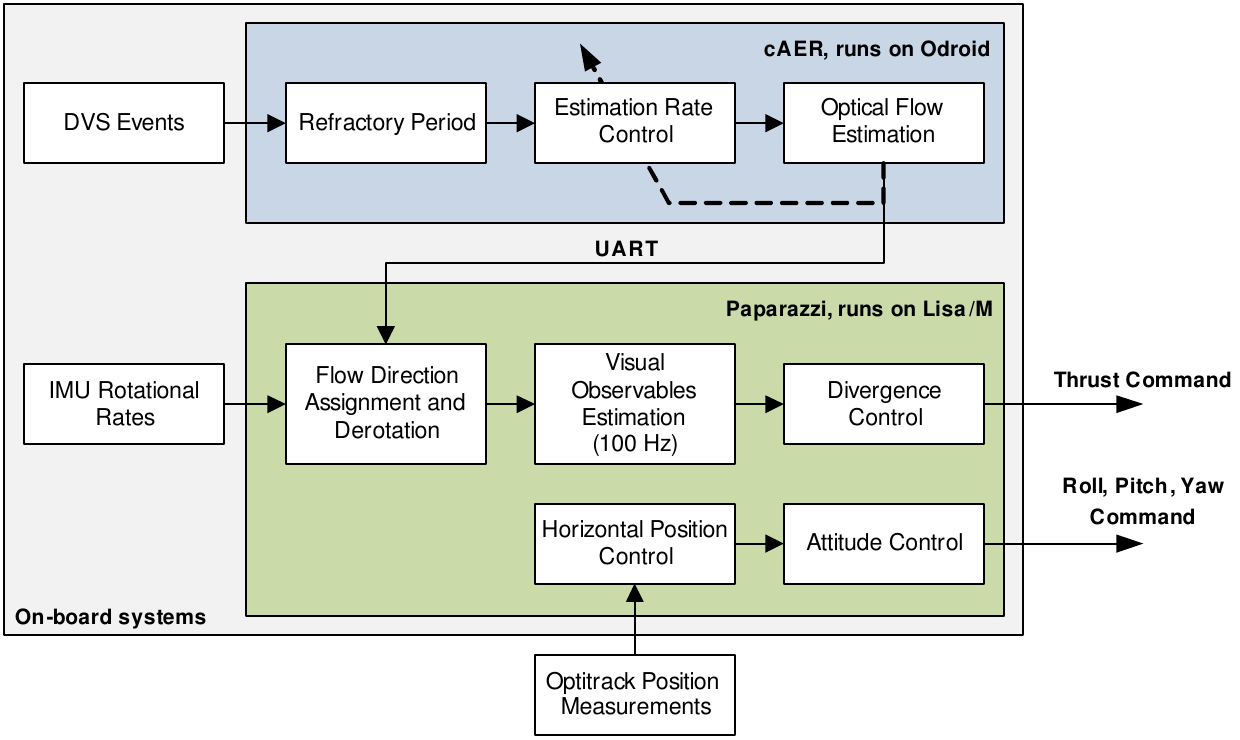}
				\label{fig:implementation}
			}
		\end{minipage}
		\caption{Overview of the experimental setup, including pictures of the MavTec. In \protect\subref{fig:mavtec_top} a top view of the vehicle is shown. The DVS is located at the bottom, protected by a foam cover. In \protect\subref{fig:mavtec_bottom} the cover is removed to expose the DVS. In \protect\subref{fig:implementation} an overview of the processing workflow is shown, indicating the distribution of processes over the Odroid and the Lisa/M processors.}
		\label{fig:mavtec}
	\end{framed}
\end{figure*}

An overview of the experimental setup is shown in \cref{fig:mavtec}, including an overview of the on-board processing workflow in \cref{fig:implementation}. The estimation pipeline is subdivided in two stages. First, raw events are transmitted from the DVS to the Odroid through a USB interface. In cAER, optical flow is computed from the events using an implementation of our optical flow algorithm. Any event for which flow is estimated, is transmitted to the Lisa/M board through a serial UART interface. This process is completely event-based and is performed in a single thread. Separate threads handle event reception and transmission through the USB and UART interfaces.

Second, in Paparazzi, a periodic follow-up processing thread runs at 100 Hz. At each iteration, all newly received optical flow events are collected and corrected for the quadrotor's attitude and rotational motion. When all new events are processed, new estimates of the scaled velocities are computed with accompanying confidence values. A separate thread running at 512 Hz performs divergence control using the new update for $\vartheta_z$, as well as horizontal position control and stabilization.


\subsection{Results}
\label{sec:results_div_landings}
Constant divergence landing maneuvers were performed for several values of the setpoint $\vartheta_{z_r}$. During the tests, the target ground location was covered with the roadmap textured mat shown in \cref{fig:roadmap}. Currently, no mechanism is implemented to account for instability of constant divergence landings at low height, as described in \citet{DeCroon2016}. Therefore, when significant self-induced oscillations are observed, the landing maneuver is manually terminated. 

\begin{figure}[h]
	\centering
	\setlength{\fwidth}{0.4\linewidth}
	\input{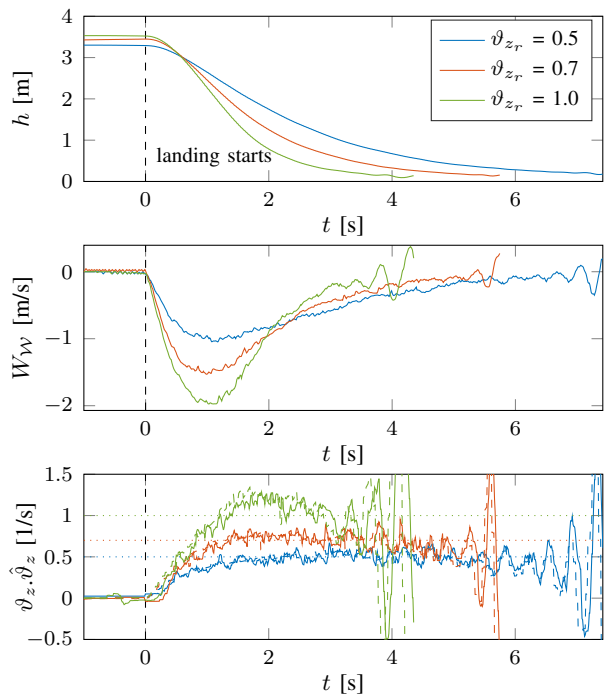}
	\caption{Height above ground, vertical speed, and divergence measurements with ground truth during a constant divergence landings performed at three different divergence setpoints. In the bottom graph, the dotted, dashed, and solid lines represent the setpoint, ground truth, and estimate for $\vartheta_z$ respectively.}
	\label{fig:const_div_landing_1}
\end{figure}

Resulting flight profiles (height, vertical speed, and divergence) are shown in \cref{fig:const_div_landing_1} for setpoints of $\vartheta_{z_r}=\lbrace0.5, 0.7,1.0\rbrace$\footnote{Video of the landings performed can be found at: \url{https://www.youtube.com/playlist?list=PL_KSX9GOn2P8RBdSyzngewi76G37PI3SF}}. Note that these values are much higher than the setpoints in comparable frame-based experiments \cite{Herisse2012,Ho2016}. The estimates for $\vartheta_z$ are shown in comparison to the ground truth estimate and the corresponding setpoint. For these maneuvers, the proportional gain $k_P$ is set to 0.2. This gain ensures that the descent remains stable during the first part. Decent tracking performance is seen for the lower two setpoints, while at $\vartheta_{z_r}=1.0$ some overshoot is observed. Still, a faster response may be obtained with an adaptive gain, such as in \citet{Ho2016a}. 

The expected instability is also clearly visible. For each setpoint, oscillations with diverging amplitude start to appear when the height is around 0.6 m above the ground, requiring the maneuver to be aborted manually at the moment. Also, a time delay is observed, whose magnitude differs between datasets. By examining the cross-correlation functions of the estimate and ground truth signals, average time delays of 0.05 s, 0.04 s, and 0.10 s are observed for the respective signals. A possible cause for this is the latency in the UART interface between the Odroid and the Lisa/M. Also, part of the delay results from the confidence filter that delays visual observable updates around zero-crossings.


In practice, the visual observable estimator thread running on the Lisa/M microprocessor does not maintain its target frequency of 100 Hz with an optical flow measurement rate $\rho_{F_{max}}$ of 2500 events per second. Instead, it drops to around 75 Hz during the landing maneuvers. However, given the limited processing power of the microprocessor, this is still a decent result. It well exceeds sampling rates seen in recent frame-based optical flow estimation pipelines, which are in the order of 15 to 25 Hz \cite{Herisse2012,Alkowatly2015,Ho2016a,DeCroon2016}. Also, with a lower setting of $\rho_{F_{max}}$ (around 2000 optical flow events per second), the target frequency of 100 Hz is well attainable. The Odroid can transmit up to approximately 8500 optical flow events per second over the UART connection, limited by the baud rate of 921.6 kB per second.

For the largest part, the maneuvers are executed successfully, even for high divergence values setpoints. With $\vartheta_{z_r}=1.0$, the MAV performs a rapid maneuver, descending from a height of 3.5 m to 1 m within 1.79 s. In comparable recent experiments with frame-based cameras for divergence measurement \cite{Ho2016a}, landings were performed up to $\vartheta_{z_r}=0.3$. Since higher values have not been attempted in these experiments, we cannot know for certain that frame-based optical flow is not applicable to such high speeds\footnote{It is planned to perform some landing tests outdoors in an unmodified outdoor environment to test the landing at even higher rates of divergence to explore the limits of our approach. The results of those tests may be included in a revised version of this paper.}.

\section{Conclusion}
\label{sec:conclusion}

In this paper we present a successful implementation of event-based optical flow estimation into a constant divergence landing controller for flying robots. Three main contributions lead to this result.

First, a novel algorithm for computing event-based optical flow is derived from an existing local plane fitting technique. The algorithm is capable of estimating normal optical flow with a wide range of magnitudes through timestamp-based clustering of the event cloud. Its performance is evaluated in ground texture scenes recorded by a DVS. Accurate estimates are seen in real event scenes with sparse, high contrast edges, as well as in scenes with densely packed, lower contrast features. Compared to the existing technique, optical flow accuracy is slightly improved for fast motion, while a larger number of successful optical flow estimates is obtained during slow motion. In addition, it is shown that the optical flow detection rate can be capped to limit computational effort for the algorithm, which enables implementation on low-end platforms without sacrificing accuracy.

Second, we introduce an algorithm for estimating optical flow based visual observables from normal optical flow measurements. By grouping flow vectors by their direction, the aperture problem can be limited by estimating the parameters of a planar optical flow field. The estimator assesses the reliability of its output through a confidence metric based on the flow estimation rate, the variance of optical flow positions, and the coefficient of determination of the flow field. When coupled to the optical flow algorithm, it is capable of estimating the visual observables accurately over a wide range of speeds. Also, the influence of fast rotational motion on the visual observables is adequately corrected through separate rotational rate measurements.

Third, using the developed pipeline, fast constant divergence landing maneuvers are demonstrated using a quadrotor equipped with a downward facing DVS. Decent tracking performance is achieved for the majority of the descent using a simple proportional controller. The final touchdown of the landing maneuver is not yet performed due to self-induced oscillations close to the ground. However, stability-based control methods have already been demonstrated that can resolve this issue. A future controller based on these methods can, for example, autonomously detect the oscillations and switch to a final touchdown phase based on constant thrust, or perform a complete landing maneuver using an adaptive gains. In addition, our controller does not yet incorporate the visual observables for horizontal stabilization, but relies on an external position tracking system. However, with the estimate accuracy and rotational motion correction presented in this work, this appears feasible.

In a first-order comparison to recent work on landing using frame-based cameras for estimating optical flow, the presented event-based pipeline demonstrates more accurate measurements at high speed and a higher sampling rate, which enable faster maneuvers than previously shown in literature. However, for a more solid conclusion regarding real-time computational benefits of event-based vision, a comparison should be performed where both frame-based and event-based cameras are incorporated in the same hardware configuration. 

While miniature frame-based bottom cameras are readily embedded into several commercially available quadrotors, the on-board hardware configuration in this work is still relatively bulky and inefficient. The implementation in this work is performed on a relatively large MAV in order to carry the weight of the DVS with separate computers for processing events and estimation of visual observables. However, the availability of smaller and lighter event-based cameras, such as the 2.2 mg meDVS, will enable high-speed optical flow control on very small flying robots. 

\appendices

%


\ifCLASSOPTIONcaptionsoff
  \newpage
\fi



\bibliographystyle{IEEEtranFix}
\bibliography{bibfile}

\begin{thebibliography}{10}
\providecommand{\url}[1]{#1}
\csname url@samestyle\endcsname
\providecommand{\newblock}{\relax}
\providecommand{\bibinfo}[2]{#2}
\providecommand{\BIBentrySTDinterwordspacing}{\spaceskip=0pt\relax}
\providecommand{\BIBentryALTinterwordstretchfactor}{4}
\providecommand{\BIBentryALTinterwordspacing}{\spaceskip=\fontdimen2\font plus
\BIBentryALTinterwordstretchfactor\fontdimen3\font minus
  \fontdimen4\font\relax}
\providecommand{\BIBforeignlanguage}[2]{{%
\expandafter\ifx\csname l@#1\endcsname\relax
\typeout{** WARNING: IEEEtran.bst: No hyphenation pattern has been}%
\typeout{** loaded for the language `#1'. Using the pattern for}%
\typeout{** the default language instead.}%
\else
\language=\csname l@#1\endcsname
\fi
#2}}
\providecommand{\BIBdecl}{\relax}
\BIBdecl

\bibitem{Floreano2015}
D.~Floreano and R.~J. Wood, ``{Science, technology and the future of small
  autonomous drones},'' \emph{Nature}, vol. 521, no. 7553, pp. 460--466, 2015.

\bibitem{DeWagter2014}
C.~{De Wagter}, S.~Tijmons, B.~D.~W. Remes, and G.~C. H.~E. {De Croon},
  ``{Autonomous flight of a 20-gram Flapping Wing MAV with a 4-gram onboard
  stereo vision system},'' in \emph{Proceedings - 2014 IEEE International
  Conference on Robotics and Automation}, 2014, pp. 4982--4987.

\bibitem{Gibson1979}
J.~J. Gibson, \emph{{The ecological approach to visual perception}}.\hskip 1em
  plus 0.5em minus 0.4em\relax Boston: Houghton Mifflin, 1979.

\bibitem{Baird2013}
E.~Baird, N.~Boeddeker, M.~R. Ibbotson, and M.~V. Srinivasan, ``{A universal
  strategy for visually guided landing},'' \emph{Proceedings of the National
  Academy of Sciences of the United States of America}, vol. 110, no.~46, pp.
  18\,686--18\,691, 2013.

\bibitem{Herisse2008}
B.~Herisse, F.~X. Russotto, T.~Hamel, and R.~Mahony, ``{Hovering flight and
  vertical landing control of a VTOL Unmanned Aerial Vehicle using optical
  flow},'' in \emph{2008 IEEE/RSJ International Conference on Intelligent
  Robots and Systems}, 2008, pp. 801--806.

\bibitem{Herisse2012}
B.~Heriss{\'{e}}, T.~Hamel, R.~Mahony, and F.-X. Russotto, ``{Landing a VTOL
  Unmanned Aerial Vehicle on a Moving Platform Using Optical Flow},''
  \emph{IEEE Transactions on Robotics}, vol.~28, no.~1, pp. 77--89, 2012.

\bibitem{Ho2016}
H.~W. Ho and G.~C. H.~E. {De Croon}, ``{Characterization of Flow Field
  Divergence for MAVs Vertical Control Landing},'' in \emph{AIAA Guidance,
  Navigation, and Control Conference}, 2016, pp. 1--13.

\bibitem{DeCroon2016}
G.~C. H.~E. {De Croon}, ``{Monocular distance estimation with optical flow
  maneuvers and efference copies: a stability-based strategy},''
  \emph{Bioinspiration {\&} Biomimetics}, vol.~11, no.~1, pp. 1--18, 2016.

\bibitem{Linares-barranco2014}
C.~Posch, T.~Serrano-Gotarredona, B.~Linares-barranco, and T.~Delbr{\"{u}}ck,
  ``{Retinomorphic Event-Based Vision Sensors : Bioinspired Cameras With
  Spiking Output},'' \emph{Proceedings of the IEEE}, vol. 102, no.~10, pp.
  1470--1484, 2014.

\bibitem{Ruffier2014}
F.~Ruffier and N.~Franceschini, ``{Optic Flow Regulation in Unsteady
  Environments: A Tethered MAV Achieves Terrain Following and Targeted Landing
  Over a Moving Platform},'' \emph{Journal of Intelligent {\&} Robotic
  Systems}, vol.~79, pp. 275--293, 2014.

\bibitem{Floreano2013}
D.~Floreano, R.~Pericet-Camara, S.~Viollet, F.~Ruffier, A.~Br{\"{u}}ckner,
  R.~Leitel, W.~Buss, M.~Menouni, F.~Expert, R.~Juston, M.~K. Dobrzynski,
  G.~L'Eplattenier, F.~Recktenwald, H.~a. Mallot, and N.~Franceschini,
  ``{Miniature curved artificial compound eyes.}'' in \emph{Proceedings of the
  National Academy of Sciences of the United States of America}, vol. 110,
  2013, pp. 9267--72.

\bibitem{Yang2015}
M.~Yang, S.-C. Liu, and T.~Delbruck, ``{A Dynamic Vision Sensor With 1{\%}
  Temporal Contrast Sensitivity and In-Pixel Asynchronous Delta Modulator for
  Event Encoding},'' \emph{IEEE Journal of Solid-State Circuits}, vol.~50,
  no.~9, pp. 2149--2160, 2015.

\bibitem{Conradt2009}
J.~Conradt, R.~Berner, M.~Cook, and T.~Delbruck, ``{An embedded AER dynamic
  vision sensor for low-latency pole balancing},'' in \emph{2009 IEEE 12th
  International Conference on Computer Vision Workshops, ICCV Workshops}, 2009,
  pp. 780--785.

\bibitem{Delbruck2013}
T.~Delbruck and M.~Lang, ``{Robotic goalie with 3 ms reaction time at 4{\%} CPU
  load using event-based dynamic vision sensor},'' \emph{Frontiers in
  Neuroscience}, vol.~7, no. 223, pp. 1--7, 2013.

\bibitem{Benosman2012}
R.~Benosman, S.-H. Ieng, C.~Clercq, C.~Bartolozzi, and M.~Srinivasan,
  ``{Asynchronous frameless event-based optical flow},'' \emph{Neural
  Networks}, vol.~27, pp. 32--37, 2012.

\bibitem{Benosman2014}
R.~Benosman, C.~Clercq, X.~Lagorce, S.-H. Ieng, and C.~Bartolozzi,
  ``{Event-based visual flow.}'' \emph{IEEE transactions on neural networks and
  learning systems}, vol.~25, no.~2, pp. 407--17, 2014.

\bibitem{Barranco2014}
F.~Barranco, C.~Fermuller, and Y.~Aloimonos, ``{Contour Motion Estimation for
  Asynchronous Event-Driven Cameras},'' \emph{Proceedings of the IEEE}, vol.
  102, no.~10, pp. 1537--1556, 2014.

\bibitem{Brosch2015}
T.~Brosch, S.~Tschechne, and H.~Neumann, ``{On event-based optical flow
  detection},'' \emph{Frontiers in Neuroscience}, vol.~9, no. 137, pp. 1--15,
  2015.

\bibitem{Bardow2016}
P.~Bardow, A.~J. Davison, and S.~Leutenegger, ``{Simultaneous Optical Flow and
  Intensity Estimation from an Event Camera},'' in \emph{Proceedings of
  Computer Vision and Pattern Recognition (CVPR)}, 2016, pp. 884--892.

\bibitem{Barranco2016}
F.~Barranco, C.~Fermuller, Y.~Aloimonos, and T.~Delbruck, ``{A Dataset for
  Visual Navigation with Neuromorphic Methods},'' \emph{Frontiers in
  Neuroscience}, vol.~10, no. February, pp. 1--9, 2016.

\bibitem{Ruckauer2016}
B.~Ruckauer and T.~Delbruck, ``{Evaluation of event-based algorithms for
  optical flow with ground-truth from inertial measurement sensor},''
  \emph{Frontiers in Neuroscience}, vol.~10, no. 176, 2016.

\bibitem{Clady2014}
X.~Clady, C.~Clercq, S.-H. Ieng, F.~Houseini, M.~Randazzo, L.~Natale,
  C.~Bartolozzi, and R.~Benosman, ``{Asynchronous visual event-based
  time-to-contact.}'' \emph{Frontiers in neuroscience}, vol.~8, no.~9, 2014.

\bibitem{Davison2007a}
A.~J. Davison, I.~D. Reid, N.~D. Molton, and O.~Stasse, ``{MonoSLAM: Real-time
  single camera SLAM},'' \emph{IEEE Transactions on Pattern Analysis and
  Machine Intelligence}, vol.~29, no.~6, pp. 1052--1067, 2007.

\bibitem{Nister2004}
D.~Nist{\'{e}}r, O.~Naroditsky, and J.~Bergen, ``{Visual odometry},'' in
  \emph{Proceedings of the 2004 IEEE Computer Society Conference on Computer
  Vision and Pattern Recognition}, vol.~1, 2004, pp. I----652.

\bibitem{Forster2014}
C.~Forster, M.~Pizzoli, and D.~Scaramuzza, ``{SVO : Fast Semi-Direct Monocular
  Visual Odometry},'' in \emph{2014 IEEE International Conference on Robotics
  and Automation (ICRA)}, 2014, pp. 15--22.

\bibitem{Engel2014}
J.~Engel, T.~Sch{\"{o}}ps, and D.~Cremers, ``{{\{}LSD-SLAM{\}}: Large-scale
  direct monocular {\{}SLAM{\}}},'' in \emph{European Conference on Computer
  Vision (ECCV)}.\hskip 1em plus 0.5em minus 0.4em\relax Springer International
  Publishing, 2014, pp. 834--849.

\bibitem{Expert2015}
F.~Expert and F.~Ruffier, ``{Flying over uneven moving terrain based on
  optic-flow cues without any need for reference frames or accelerometers.}''
  \emph{Bioinspiration {\&} biomimetics}, vol.~10, 2015.

\bibitem{Srinivasan1996}
M.~Srinivasan, S.~Zhang, M.~Lehrer, and T.~Collett, ``{Honeybee navigation en
  route to the goal: visual flight control and odometry},'' \emph{Journal of
  Experimental Biology}, vol. 199, no.~1, pp. 237--244, 1996.

\bibitem{Alkowatly2015}
M.~T. Alkowatly, V.~M. Becerra, and W.~Holderbaum, ``{Bioinspired Autonomous
  Visual Vertical Control of a Quadrotor Unmanned Aerial Vehicle},''
  \emph{Journal of Guidance, Control, and Dynamics}, vol.~38, no.~2, pp.
  249--262, 2015.

\bibitem{Lee1976}
D.~N. Lee, ``{A theory of visual control of braking based on information about
  time-to-collision.}'' \emph{Perception}, vol.~5, no.~4, pp. 437--459, 1976.

\bibitem{Ho2016a}
\BIBentryALTinterwordspacing
H.~W. Ho, G.~C. H.~E. de~Croon, E.~van Kampen, Q.~P. Chu, and M.~Mulder,
  ``{Adaptive Control Strategy for Constant Optical Flow Divergence Landing},''
  pp. 1--14, 2016. [Online]. Available: \url{http://arxiv.org/abs/1609.06767}
\BIBentrySTDinterwordspacing

\bibitem{Lichtsteiner2008}
P.~Lichtsteiner, C.~Posch, and T.~Delbruck, ``{A 128x128 120 dB 15 $\mu$s
  Latency Asynchronous Temporal Contrast Vision Sensor},'' \emph{IEEE Journal
  of Solid-State Circuits}, vol.~43, no.~2, pp. 566--576, 2008.

\bibitem{Cho2015}
D.-i.~D. Cho and T.-j. Lee, ``{A Review of Bioinspired Vision Sensors and Their
  Applications},'' \emph{Sensors and Materials}, vol.~27, no.~6, pp. 447--463,
  2015.

\bibitem{Posch2011}
C.~Posch, D.~Matolin, and R.~Wohlgenannt, ``{A QVGA 143 dB dynamic range
  frame-free PWM image sensor with lossless pixel-level video compression and
  time-domain CDS},'' \emph{IEEE Journal of Solid-State Circuits}, vol.~46,
  no.~1, pp. 259--275, 2011.

\bibitem{Brandli2014}
C.~Brandli, R.~Berner, M.~Yang, S.-C. Liu, and T.~Delbruck, ``{A 240x180 130 dB
  3 $\mu$s Latency Global Shutter Spatiotemporal Vision Sensor},'' \emph{IEEE
  Journal of Solid-State Circuits}, vol.~49, no.~10, pp. 2333--2341, 2014.

\bibitem{Conradt2015}
J.~Conradt, ``{On-Board Real-Time Optic-Flow for Miniature Event-Based Vision
  Sensors},'' in \emph{2015 IEEE International Conference on Robotics and
  Biomimetics (ROBIO)}, Zhuhai, China, 2015.

\bibitem{Weikersdorfer2013}
D.~Weikersdorfer, R.~Hoffmann, and J.~Conradt, ``{Simultaneous localization and
  mapping for event-based vision systems},'' in \emph{International Conference
  on Computer Vision Systems}.\hskip 1em plus 0.5em minus 0.4em\relax
  Springer-Verlag Berlin Heidelberg, 2013, pp. 133--142.

\bibitem{Weikersdorfer2014}
D.~Weikersdorfer, D.~B. Adrian, and D.~Cremers, ``{Event-based 3D SLAM with a
  depth-augmented dynamic vision sensor},'' \emph{2014 IEEE International
  Conference on Robotics and Automation (ICRA)}, pp. 359--364, 2014.

\bibitem{Mueggler2014}
E.~Mueggler, B.~Huber, and D.~Scaramuzza, ``{Event-based, 6-DOF Pose Tracking
  for High-Speed Maneuvers},'' in \emph{2014 IEEE/RSJ International Conference
  on Intelligent Robots and Systems}, 2014, pp. 2761--2768.

\bibitem{Kim2014}
H.~Kim, ``{Simultaneous Mosaicing and Tracking with an Event Camera},'' in
  \emph{Proceedings of the British Machine Vision Conference 2014}, 2014.

\bibitem{Kim2016}
H.~Kim, S.~Leutenegger, and A.~J. Davison, ``{Real-Time 3D Reconstruction and
  6-DoF Tracking with an Event Camera},'' in \emph{European Conference on
  Computer Vision}.\hskip 1em plus 0.5em minus 0.4em\relax Amsterdam: Springer
  International Publishing, 2016, pp. 349--364.

\bibitem{Kueng2016}
B.~Kueng, E.~Mueggler, G.~Gallego, and D.~Scaramuzza, ``{Low-Latency Visual
  Odometry using Event-based Feature Tracks},'' in \emph{IEEE/RSJ International
  Conference on Intelligent Robots and Systems (IROS)}, 2016.

\bibitem{Baker2011}
S.~Baker, D.~Scharstein, J.~P. Lewis, S.~Roth, M.~J. Black, and R.~Szeliski,
  ``{A database and evaluation methodology for optical flow},''
  \emph{International Journal of Computer Vision}, vol.~92, no.~1, pp. 1--31,
  2011.

\bibitem{Horn1981}
B.~K. Horn and B.~G. Schunck, ``{Determining optical flow},'' \emph{Artificial
  Intelligence}, vol.~17, pp. 185--203, 1981.

\bibitem{Lucas1981}
B.~D. Lucas and T.~Kanade, ``{An iterative image registration technique with an
  application to stereo vision},'' in \emph{International Joint Conference on
  Artificial Intelligence}, vol.~81, 1981, pp. 674--679.

\bibitem{Beauchemin1995}
S.~S. Beauchemin and J.~L. Barron, ``{The computation of optical flow},''
  \emph{ACM Computing Surveys}, vol.~27, no.~3, pp. 433--466, 1995.

\bibitem{Rosten2008}
\BIBentryALTinterwordspacing
E.~Rosten, R.~Porter, and T.~Drummond, ``{Faster and better: a machine learning
  approach to corner detection},'' \emph{IEEE Transactions on Pattern Analysis
  and Machine Intelligence}, vol.~32, no.~1, pp. 105--19, 2008. [Online].
  Available: \url{http://arxiv.org/pdf/0810.2434}
\BIBentrySTDinterwordspacing

\bibitem{DeCroon2013}
G.~C. H.~E. {De Croon}, H.~W. Ho, C.~{De Wagter}, E.~{Van Kampen}, B.~Remes,
  and Q.~P. Chu, ``{Optic-flow based slope estimation for autonomous
  landing},'' \emph{International Journal of Micro Air Vehicles}, vol.~5,
  no.~4, pp. 287--297, 2013.

\bibitem{Bouguet2000}
\BIBentryALTinterwordspacing
J.-Y. Bouguet, ``{Pyramidal implementation of the Lucas Kanade feature tracker
  - Description of the Algorithm},'' \emph{Intel Corporation, Microprocessor
  Research Labs}, vol.~5, 2001. [Online]. Available:
  \url{http://robots.stanford.edu/cs223b04/algo\_affine\_tracking.pdf}
\BIBentrySTDinterwordspacing

\bibitem{Zufferey2010}
J.-C. Zufferey, A.~Beyeler, and D.~Floreano, ``{Autonomous flight at low
  altitude using light sensors and little computational power},''
  \emph{International Journal of Micro Air Vehicles}, vol.~2, no.~2, pp.
  107--117, 2010.

\bibitem{Ruffier2005}
F.~Ruffier and N.~Franceschini, ``{Optic flow regulation: The key to aircraft
  automatic guidance},'' \emph{Robotics and Autonomous Systems}, vol.~50,
  no.~4, pp. 177--194, 2005.

\bibitem{Clady2015}
X.~Clady, S.-H. Ieng, and R.~Benosman, ``{Asynchronous event-based corner
  detection and matching.}'' \emph{Neural Networks}, vol.~66, pp. 91--106,
  2015.

\bibitem{Barranco2015}
F.~Barranco, C.~Fermuller, and Y.~Aloimonos, ``{Bio-inspired Motion Estimation
  with Event-Driven Sensors},'' in \emph{Advances in Computational
  Intelligence}.\hskip 1em plus 0.5em minus 0.4em\relax Springer International
  Publishing, 2015, pp. 309--321.

\bibitem{Longuet-Higgins1980}
H.~C. Longuet-Higgins and K.~Prazdny, ``{The interpretation of a moving retinal
  image.}'' \emph{Proceedings of the Royal Society of London, B: Biological
  Sciences}, vol. 208, no. 1173, pp. 385--397, 1980.

\bibitem{Brown1966}
D.~Brown, ``{Decentering Distortion of Lenses},'' \emph{Photometric
  Engineering}, vol.~32, no.~3, pp. 444--462, 1966.

\bibitem{Bouguet1999}
\BIBentryALTinterwordspacing
J.-Y. Bouguet, ``{Complete Camera Calibration Toolbox for Matlab},'' 1999.
  [Online]. Available: \url{http://www.vision.caltech.edu/bouguetj/calib\_doc/}
\BIBentrySTDinterwordspacing

\bibitem{Heath2002}
M.~T. Heath, \emph{{Scientific computing: an introductory survey}},
  2nd~ed.\hskip 1em plus 0.5em minus 0.4em\relax New York: McGraw-Hill, 2002.

\bibitem{Weisberg2005}
S.~Weisberg, \emph{{Applied linear regression}}.\hskip 1em plus 0.5em minus
  0.4em\relax John Wiley {\&} Sons, 2005.

\bibitem{Longinotti2014}
\BIBentryALTinterwordspacing
L.~Longinotti, ``{cAER: A framework for event-based processing on embedded
  systems},'' BSc Thesis, University of Z{\"{u}}rich, 2014. [Online].
  Available: \url{http://sourceforge.net/projects/jaer/files/cAER/}
\BIBentrySTDinterwordspacing

\end{thebibliography}
\end{document}